\pdfoutput=1

\documentclass[11pt]{article}

\usepackage{ACL2023}

\usepackage{times}
\usepackage{latexsym}

\usepackage[T1]{fontenc}

\usepackage[utf8]{inputenc}

\usepackage{microtype}

\usepackage{inconsolata}

\usepackage{tikz}

\usepackage{array}
\usepackage{makecell}

\usepackage{amsmath}
\usepackage{amsfonts}
\usepackage{amssymb}
\usepackage{graphicx}
\usepackage{geometry}
\usepackage{natbib}
\usepackage{hyperref}
\usepackage{booktabs}
\usepackage{algorithm}
\usepackage{algorithmic}
\usepackage{multirow}
\usepackage{adjustbox}  
\usepackage[most]{tcolorbox}
\tcbuselibrary{listings,breakable}

\usepackage{pifont, xcolor}
\newcommand{\cmark}{\textcolor{green}{\ding{51}}}
\newcommand{\xmark}{\textcolor{red}{\ding{55}}}

\newcommand{\ds}{DocSplit}
\newcommand{\dsPolySeq}{DocSplit-Poly-Seq}
\newcommand{\dsPolyInt}{DocSplit-Poly-Int}
\newcommand{\dsPolyRand}{DocSplit-Poly-Rand}
\newcommand{\dsMonoSeq}{DocSplit-Mono-Seq}
\newcommand{\dsMonoRand}{DocSplit-Mono-Rand}

\usepackage{tcolorbox}
\tcbuselibrary{skins} 
\definecolor{todo-yellow}{RGB}{255,253,160}
\definecolor{todo-border}{RGB}{200,200,200} 
\definecolor{todo-text}{RGB}{0,0,150} 

\title{DocSplit: A Comprehensive Benchmark Dataset and Evaluation Approach for Document Packet Recognition and Splitting}

\author{\textbf{Md Mofijul Islam, Md Sirajus Salekin, Nivedha Balakrishnan, Vincil C. Bishop III,}\\ \textbf{Niharika Jain, Spencer Romo, Bob Strahan, Boyi Xie, Diego A. Socolinsky} \\ \textbf{Amazon Web Services}}

\begin{document}

\maketitle

\begin{abstract} 
Document understanding in real-world applications often requires processing heterogeneous, multi-page document packets containing multiple documents stitched together. Despite recent advances in visual document understanding, the fundamental task of document packet splitting, which involves separating a document packet into individual units, remains largely unaddressed. We present the first comprehensive benchmark dataset, $\textit{DocSplit}$, along with novel evaluation metrics for assessing the document packet splitting capabilities of large language models. $\textit{DocSplit}$ comprises five datasets of varying complexity, covering diverse document types, layouts, and multimodal settings. We formalize the $\textit{DocSplit}$ task, which requires models to identify document boundaries, classify document types, and maintain correct page ordering within a document packet. The benchmark addresses real-world challenges, including out-of-order pages, interleaved documents, and documents lacking clear demarcations. We conduct extensive experiments evaluating multimodal LLMs on our datasets, revealing significant performance gaps in current models' ability to handle complex document splitting tasks. The $\textit{DocSplit}$ benchmark datasets and proposed novel evaluation metrics provide a systematic framework for advancing document understanding capabilities essential for legal, financial, healthcare, and other document-intensive domains. We release the datasets\footnote{https://huggingface.co/datasets/amazon/doc\_split \label{fn:dataset}} to facilitate future research in document packet processing.
\end{abstract}

\section{Introduction}
\begin{figure}[!t]
    \centering
    \includegraphics[width=\columnwidth]{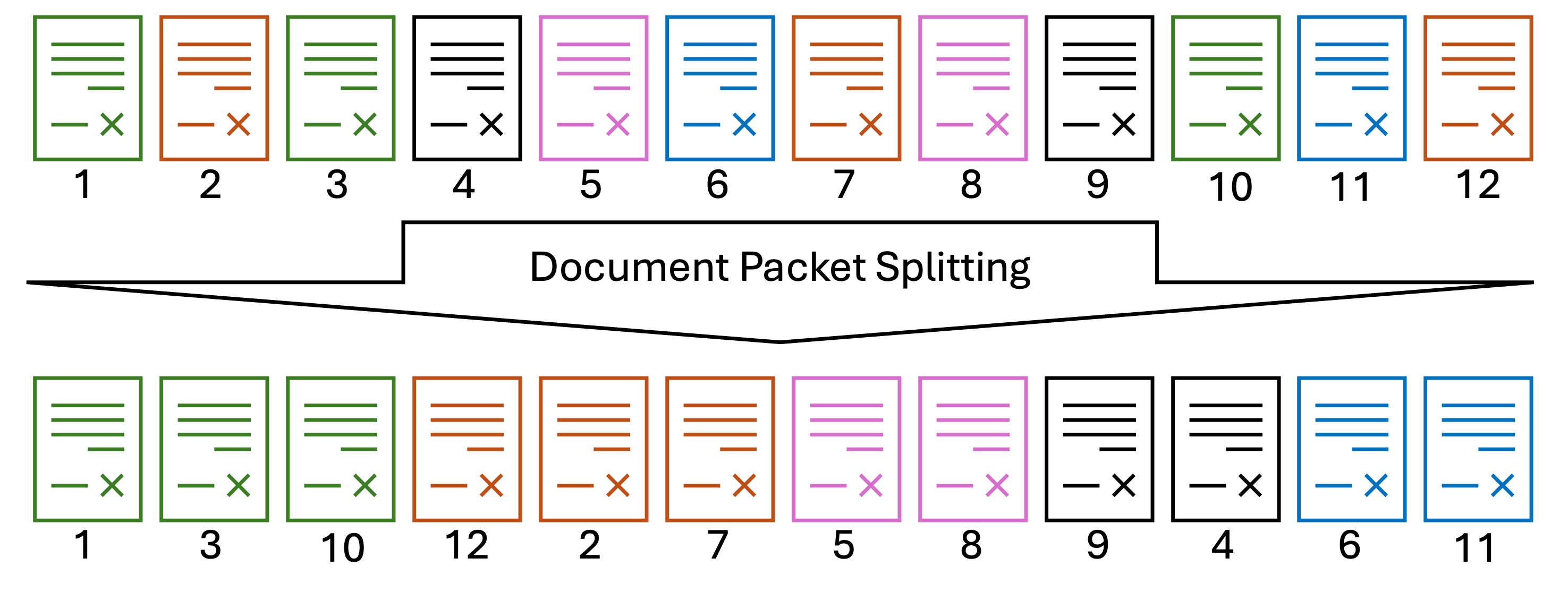}
    \caption{The document packet splitting task, where an unordered sequence of pages from multiple documents (shown at top) must be correctly grouped and reordered into their constituent documents (shown at bottom). The colors represent individual documents, while the numbers indicate the original page indices from the top input sequence, demonstrating how pages from the same document can be shuffled and interleaved in the input packet.}
    \label{fig:enter-label}
\end{figure}

In many real-world applications, document processing faces a critical challenge: organizing a collection of loosely ordered document sequence received as a packet (Figure~\ref{fig:enter-label}) \cite{wang2025document,VanLandeghem2024}. 
While the field has achieved remarkable progress in single-page document classification \cite{harley2015evaluation,luo2025bivldocbidirectionalvisionlanguagemodeling} and visual document understanding systems \cite{Duan2025,Chen2025}, a fundamental challenge remains largely underexplored: the decomposition of document packets into their constituent logical components.

This challenge transcends simple document categorization, demanding reasoning about document boundaries, content relationships, and structural coherence in a document packet \cite{li2025docsam}. The problem manifests in bundled submissions without clear separators, manual assembly errors, or legacy collections with lost structure. The complexity intensifies when pages from individual documents become dispersed, requiring models to simultaneously perform boundary detection, type classification, and sequence reconstruction.

The practical implications span numerous high-stakes domains where document processing accuracy directly impacts operational efficiency and decision-making quality. Financial institutions, legal firms, and healthcare systems must extract and organize documents from mixed submissions such as loan applications with supporting documentation, case discovery evidence, and insurance claims with medical records, each presenting unique challenges in document variety, formatting inconsistencies, and quality variations.

Current approaches to document understanding fall short of addressing these challenges. Existing benchmarks like RVL-CDIP \cite{harley2015evaluation} focus exclusively on single-page classification, while recent multimodal large language models \cite{Wang2025marten,Zhang2024dockylin,Liao2024doclayllm} have not been systematically evaluated on document packet splitting. The few attempts to address multi-page scenarios \cite{VanLandeghem2024,wang2025document} lack comprehensive benchmark datasets capturing real-world packet splitting challenges. Recent work on multi-document processing \cite{lior2024seam} has highlighted the need for systematic evaluation frameworks, yet none address document boundary detection within concatenated packets.

To address this gap, we introduce \textbf{DocSplit} (\textbf{Doc}ument Packet \textbf{Split}ting), the first comprehensive benchmark for document packet splitting. We formalize the \textit{DocSplit} task as transforming an input sequence of $N$ pages from a document packet into structured representations that capture both document boundaries and classifications. {\ds} comprises five carefully curated datasets of varying complexity, spanning diverse document types, layouts, and multimodal settings reflecting real-world document processing scenarios.
Our contributions are multifaceted and fundamental:
\begin{itemize}
\item \textbf{Introduce the first comprehensive benchmark} for document packet splitting tasks.
\item \textbf{Define the DocSplit task} with a formal definition for boundary detection, classification, and page ordering.
\item \textbf{Develop and release five benchmark datasets} with varying complexity, covering diverse document types, shuffled pages, interleaved documents, and multimodal settings.
\item \textbf{Design the first evaluation framework} with novel metrics for boundary detection, classification, and page sequence reconstruction.
\item \textbf{Benchmark state-of-the-art LLMs}, exposing substantial performance gaps on shuffled and interleaved documents that underscore the necessity for domain-specific models.
\end{itemize}


\section{Real-world Use Cases} 
Document packet splitting manifests across multiple high-stakes industries where manual assembly creates scrambled documents. For example, in \textbf{healthcare}, medical claims processing involves packets containing prescription records, laboratory results, physician notes, and insurance forms from disparate systems with varying formats and quality. Assembly processes cause page duplication, combined patient records, and omitted documentation, requiring dedicated specialists for accurate extraction and claim adjudication. In \textbf{transportation and logistics}, proof-of-delivery packets that contain rate confirmations and bills of lading photographed by drivers at delivery sites, causing page shuffling and poor image quality. In \textbf{mortgage} applications, property transaction packets contain deeds, mortgages, liens, and tax records from multiple sources with inconsistent formats, where pages may be scanned randomly with duplicates or omissions, impacting verification efficiency and transaction costs. These challenges across healthcare, insurance, banking, finance, and energy sectors motivate us to build robust benchmarking solutions for automated packet splitting.

\section{Related Work}
Document classification has evolved from early CNN \cite{harley2015icdar} through transformer architectures \cite{larson2023evaluationdocumentclassificationusing} to state-of-the-art multimodal LLMs \cite{luo2025bivldocbidirectionalvisionlanguagemodeling}, with various single-page classification benchmarks \cite{harley2015evaluation}. However, real-world use-cases typically require the ability to classify across many pages. Van~Landeghem et al.~\cite{VanLandeghem2024} proposed tasks such as page stream classification and bundle classification, motivating the document splitting task i.e. segmenting concatenated documents into individual units. 

Recent Visual Document Understanding (VDU) benchmarks \cite{Ouyang2025,Chen2025,Guo2025,Caffagni2025} focus on retrieval and comprehension, leaving document segmentation largely unaddressed. Synthetic data generation \cite{Jiang2025,Duan2025} and layout understanding approaches \cite{Luo2024,Zhu2025layouttoken} have enhanced document processing, while specialized multimodal LLMs \cite{Wang2025marten,Zhang2024dockylin,Liao2024doclayllm} have emerged for document-specific tasks.

Despite this progress, the document splitting task remains underexplored. While Van Landeghem et al.~\cite{VanLandeghem2024} recognized and formally defined the document splitting task, they did not propose any benchmark dataset covering the various real-world variations of document splitting scenarios. Our work fills this critical gap by being the first to propose five different benchmark datasets with varying complexity, comprehensively covering diverse document splitting tasks and novel evaluation metrics. These benchmarks enable systematic evaluation of methods for accurately detecting inter-document boundaries across different document types, layouts, and multimodal settings.
\section{Document Packet Splitting Task}

The multi-page document packet classification and splitting (\textit{DocSplit}) task involves the grouping, classification, and ordering of a document packet. 

The \textit{DocSplit} task aims to transform an input sequence of $N$ pages from a document packet into a structured representation that captures both the document boundaries and the inherent categorization of each document. We denote the packet as a set of pages $P = \{p_1, p_2, \cdots, p_n\}$. The primary objectives are twofold: first, to identify document boundaries $B = \{(s_1, e_1, t_1), (s_2, e_2, t_2), \cdots, (s_k, e_k, t_k)\}$, where $s_i$ and $e_i$ represent start and end page indices, and $t_i$ denotes the document type; and second, to determine the correct sequential ordering $O = \{o_1, o_2, \ldots, o_n\}$ of pages within each identified document.

\textbf{Evaluation Challenges}: Document packet splitting encompasses two critical and interdependent sub-problems: \textbf{document clustering} and \textbf{page ordering}. Document clustering involves correctly grouping pages that belong to the same logical document, while page ordering requires accurately determining the sequential arrangement of pages within each document.

\textbf{Mathematical Formulation}: Document clustering can be formalized as a partition problem where the algorithm must identify the correct grouping of pages into distinct documents. The ground truth partition $\mathcal{D} = \{D_1, D_2, \ldots, D_m\}$ divides $P$ into $m$ disjoint subsets, where each subset $D_i$ represents a distinct document. Similarly, the predicted partition $\hat{\mathcal{D}} = \{\hat{D}_1, \hat{D}_2, \ldots, \hat{D}_k\}$ represents the algorithm's grouping of pages. The quality of clustering can be evaluated by comparing the similarity between $\mathcal{D}$ and $\hat{\mathcal{D}}$. Within each document, page ordering can be conceptualized as a ranking problem. For a document $D_i$ containing $|D_i|$ pages, the ground truth ordering is a permutation $\sigma_i: D_i \rightarrow \{1, 2, \ldots, |D_i|\}$ that maps each page to its correct position in the sequence. Similarly, the predicted ordering is a permutation $\hat{\sigma}_i: D_i \rightarrow \{1, 2, \ldots, |D_i|\}$. The quality of ordering can be evaluated by measuring the correlation between these two permutations.

\section{Proposed Evaluation Framework}

\textbf{Clustering Performance Metrics}: To evaluate clustering performance, we employ the Rand Index (RI) \cite{hubert1985comparing}, which measures similarity between two clusterings by considering all pairs of elements: $\text{RI} = \frac{a + b}{\binom{n}{2}}$,
where $a$ and $b$ count pairs consistently assigned in both $\mathcal{D}$ and $\hat{\mathcal{D}}$, and $n$ is the number of pages in $P$. The Rand Index ranges from 0 to 1, with 1 indicating perfect agreement. For document packet splitting, we additionally employ the V-measure metric, defined as the harmonic mean of homogeneity and completeness: $V = (2 \cdot h \cdot c) / (h + c)$, where $h$ (homogeneity) measures whether each predicted cluster contains only members of a single ground truth class, and $c$ (completeness) measures whether all members of a ground truth class are assigned to the same predicted cluster. V-measure ranges from 0 to 1, with 1 indicating perfect clustering.
To capture different aspects of clustering quality, we define a weighted combination: $S_{\text{clustering}} = w \cdot V + (1-w) \cdot \text{RI}$, where $w \in [0,1]$ controls the relative importance of V-measure versus Rand Index.


\begin{table*}[!ht]
\small
	    \centering 
    \scalebox{0.95}{
    \makebox[\textwidth]{
    \centering 
    \begin{tabular}{@{}lcccccc@{}}
        \toprule
        \textbf{Dataset} &
        \textbf{Purpose} &
        $\mathbf{\#d}$ &
		$\mathbf{\#p}$ &
		$\mathbf{|\mathcal{Y}|}$ &
        \textbf{Language} &
        \textbf{Color depth} \\ \midrule
    NIST~\cite{dimmick1992nist}             & $f_s$                            &  -   & 5590                  & 20 & English       & Grayscale \\
    MARG~\cite{long2005image}            & $f_s$                            &   -  & 1553                  & 2 & English       & RGB \\
        Tobacco-800~\cite{zhu2007automatic}             & $f_s$                            &  -   & 800                  & 2 & English       & Grayscale \\
         TAB~\cite{mungmeeprued2022tab}             & $f_s$                            &  -   & 44.8K                  & 2 & English       & Grayscale \\
  Tobacco-3482~\cite{kumar2013unsupervised}             & $f_p$                            &  -   & 3482                  & 10 & English       & Grayscale \\
        RVL-CDIP~\cite{harley2015evaluation}                 & pre-training, $f_p$ &   -  & 400K                  & 16 & English       & Grayscale \\
        RVL-CDIP-N\_MP \cite{van2024beyond}  & $f_d$, OOD                             & 1002   & $\mathbb{E}[L]=10$ & 16 & English       & RGB       \\
        \textbf{{\ds}} & $f_{dp}$, $f_p$, $f_d$, OOD &52.6K &~1.55M &13 & Mixed &
        Mixed \\ 
        \bottomrule
    \end{tabular}
}}
\caption{Comparative statistics of single-page and multi-page document classification datasets. We present dataset characteristics including intended purpose ($f_{dp}$: document packet splitting, $f_s$: form segmentation, $f_p$: form processing, $f_d$: form detection), corpus size ($\#d$: documents, $\#p$: pages), label space cardinality ($|Y|$), language coverage, and image specifications. Multi-page (MP) variants report expected document length $E[L]$. OOD indicates datasets curated for out-of-distribution evaluation. \textbf{Detailed dataset comparison is include in the Appendix}.}
\label{tab:dc}
\end{table*}

\textbf{Page Ordering Performance Metrics}: To measure page ordering quality within individual documents, we employ Kendall's Tau coefficient ($\tau$) \cite{kendall1938new}, which measures ordinal association between two rankings: $\tau = (n_c - n_d) / \binom{|D_i|}{2}$, where $n_c$ and $n_d$ count concordant and discordant pairs, and $|D_i|$ is the number of pages within document $D_i$. Kendall's Tau ranges from -1 to 1, where $\tau = 1$ indicates perfect positive correlation, $\tau = 0$ indicates no correlation, and $\tau = -1$ indicates perfect negative correlation.
We compute the average Kendall's Tau across all multi-page documents: $S_{\text{ordering}} = \frac{1}{|\mathcal{M}|} \sum_{i \in \mathcal{M}} \tau_i$, where $\mathcal{M} = \{i : |D_i| > 1\}$ represents the set of documents with more than one page, and $\tau_i$ measures how well the predicted ordering $\hat{\sigma}_i$ aligns with ground truth ordering $\sigma_i$ for document $D_i$.


\textbf{Combined Evaluation Score}: To evaluate overall performance, we propose a combined score integrating both clustering and ordering metrics: $S_{\text{packet}} = \alpha \cdot S_{\text{clustering}}  + \beta \cdot S_{\text{ordering}}$, where $\alpha \geq 0$ and $\beta \geq 0$ control the weights of clustering and ordering performance ($\alpha+\beta = 1$). The inclusion of V-measure alongside the combined clustering score provides additional emphasis on homogeneity and completeness, critical for packet splitting task.


\textbf{Metrics Rationale}: We combine Rand Index and V-measure for comprehensive clustering evaluation. Rand Index quantifies pairwise classification accuracy, while V-measure ensures homogeneity (single-class clusters) and completeness (no fragmentation). 
Kendall's Tau measures pairwise ordering correlation, focusing on relative sequence rather than absolute positions. The normalized range [-1, 1] allows the comparison between documents of varying lengths.

\textbf{Framework Flexibility}: The framework offers flexibility through parameters $(w, \alpha, \beta)$ to emphasize clustering or ordering based on task priorities. With default parameters $\alpha = \beta = 0.5$ and $w = 0.5$, the theoretical range spans from $-0.5$ to $1$, where $S_{\text{packet}} = 1.0$ and $S_{\text{packet}} = -0.5 $ indicate best performance and worst performance respectively. It is worth noting that both clustering or ordering metrics can be substituted by any other potential metrics as long as they are relevant to the use case.

\section{DocSplit Benchmark Dataset}
\label{sec:dataset_curation}

\textbf{Data Source and Preprocessing}: Our benchmark datasets\footref{fn:dataset} are derived from the RVL-CDIP-MP dataset \cite{VanLandeghem2024}, a multi-page extension of RVL-CDIP \cite{harley2015icdar}. We removed 3 corrupted files from the original 998 documents, yielding 995 valid documents comprising 5,231 pages across 13 categories. Document preprocessing involved converting each page to image format and extracting text using Amazon Textract and deepseek OCR. Both representations were stored in HuggingFace-compatible format with metadata for downstream benchmark creation. We performed category-wise stratified splitting with 55\% training, 20\% validation, and 25\% testing to prevent data leakage, ensuring no document appears across multiple splits.

\textbf{Benchmark Dataset Variations}: We created five distinct benchmarks targeting different document concatenation scenarios. These benchmarks evaluate models' abilities to identify document boundaries, classify document types, and maintain correct page ordering.

\textbf{a) {\dsMonoSeq} (Single Category Document Concatenation Sequentially):} Creates document packets by randomly selecting a single document category (excluding 'language') and concatenating documents until reaching the target page count. This benchmark addresses scenarios requiring boundary detection without category transitions as discriminative signals, such as legal document processing where multiple contracts of the same type are bundled together.

\textbf{b) {\dsMonoRand} (Single Category Document Pages Randomization):} Similar to {\dsMonoSeq}, but shuffles pages from the concatenated documents. This benchmark evaluates models' robustness to page-level disruptions common in manual document assembly, requiring boundary detection and page sequence reconstruction.

\textbf{c) {\dsPolySeq} (Multi Category Documents Concatenation Sequentially):} Creates document packets by concatenating documents from different categories without repetition while preserving page ordering. This benchmark simulates heterogeneous document assembly scenarios such as medical claims processing, testing models' ability to detect inter-document boundaries.

\textbf{d) {\dsPolyInt} (Multi Category Document Pages Interleaving):} Similar to {\dsPolySeq}, but interleaves pages in a round-robin fashion. This benchmark simulates batch processing scenarios such as mortgage processing where property deeds, tax records, and legal notices are interspersed, challenging models to identify which non-contiguous pages belong together while preserving ordering within each document.

\textbf{e) {\dsPolyRand} (Multi Category Document Pages Randomization):} Similar to {\dsPolySeq}, but applies complete randomization across all pages, representing maximum entropy scenarios. This stress-tests model robustness under worst-case conditions where no structural assumptions hold, such as document management system failures or emergency document recovery, requiring models to rely solely on content-based features for clustering and sequence reconstruction.

\textbf{Ground Truth Annotation}: Each benchmark sample includes structured ground truth annotations comprising document metadata, page-level identifiers, and sequencing information, enabling comprehensive evaluation of packet splitting performance through our proposed metrics to assess both document boundary detection and page sequence reconstruction (detailed annotation schema provided in Appendix).

\section{Experimental Setup}
We evaluated five {\ds} benchmark variations (Section~\ref{sec:dataset_curation}) using Claude Sonnet 4.5, Claude Haiku 4.5, DeepSeek, Gemma, and Qwen. All models were configured with temperature 0.0, top-p 0.1, top-k 5, and maximum output of 4,096 tokens to accommodate structured JSON responses containing boundary predictions, type classifications, and page ordinal assignments.

We employed two evaluation approaches. First, our composite scoring framework combines clustering metrics (V-measure and Rand Index, $w=0.5$) with ordering metrics (Kendall's Tau), weighted equally ($\alpha=\beta=0.5$). Second, we measured three page-level accuracy metrics (classical) with progressively stricter criteria: \textbf{Page Level Accuracy} measures whether the system correctly predicts the document type. \textbf{Page+Split Accuracy} measures whether the system correctly predicts both the document type and document group ID. \textbf{Page+Split+Order Accuracy} measures whether the system correctly predicts the document type, document group ID, and page ordinal..

\section{Results and Discussion}
\begin{table*}[t]
\centering
\small
\setlength{\tabcolsep}{3.5pt}
\begin{tabular}{@{}cl ccc c cc@{}}
\toprule
& & \multicolumn{3}{c}{\textbf{Proposed}} & & \multicolumn{2}{c}{\textbf{Classical}} \\
\cmidrule(lr){3-5} \cmidrule(lr){7-8}
\textbf{Benchmark} & \textbf{Model} & \textbf{Clustering} & \textbf{Ordering} & \textbf{Packet} & \textbf{Page} & \textbf{Page+Split} & \textbf{Page+Split+Order} \\
\midrule
\multirow{7}{*}{\textbf{DocSplit-Mono-Seq}} 
& Claude Haiku 4.5   & 0.8433 & 0.9950 & 0.9191 & 0.7919 & 0.7326 & 0.7316 \\
& Claude Sonnet 4.5  & 0.8542 & \textbf{0.9960} & 0.9251 & 0.8239 & 0.7656 & 0.7656 \\
& DeepSeek V3.1      & 0.8473 & 0.9953 & 0.9213 & 0.8242 & 0.6831 & 0.6821 \\
& Gemma 3 27B        & 0.6093 & 0.9940 & 0.8016 & 0.7257 & 0.3134 & 0.3134 \\
& Qwen 3 VL 235B& \textbf{0.8831} & \textbf{0.9960} & \textbf{0.9395} & \textbf{0.8418} & \textbf{0.8036} & \textbf{0.8026} \\
\midrule
\multirow{7}{*}{\textbf{DocSplit-Mono-Rand}} 
& Claude Haiku 4.5   & 0.8138 & 0.9963 & 0.9051 & 0.7742 & 0.6777 & 0.6762 \\
& Claude Sonnet 4.5  & 0.8391 & \textbf{1.0000 }& 0.9195 & 0.8312 & 0.7229 & 0.7229 \\
& DeepSeek V3.1      & 0.8072 & \textbf{1.0000} & 0.9036 & 0.7957 & 0.6265 & 0.6262 \\
& Gemma 3 27B        & 0.5584 & 0.9980 & 0.7782 & 0.7240 & 0.2601 & 0.2601 \\
& Qwen 3 VL 235B& \textbf{0.8611} & 0.9980 & \textbf{0.9295} & \textbf{0.8362} & \textbf{0.7611} & \textbf{0.7611} \\
\midrule
\multirow{7}{*}{\textbf{DocSplit-Poly-Seq}} 
& Claude Haiku 4.5   & 0.8811 & 0.9971 & 0.9391 & 0.7930 & 0.7594 & 0.7576 \\
& Claude Sonnet 4.5  & 0.8755 & \textbf{1.0000} & 0.9377 & 0.7910 & 0.7593 & 0.7593 \\
& DeepSeek V3.1      & 0.8898 & 0.9971 & 0.9435 & 0.7961 & 0.7221 & 0.7215 \\
& Gemma 3 27B        & 0.7868 & 0.9980 & 0.8924 & 0.5329 & 0.3107 & 0.3107 \\
& Qwen 3 VL 235B& \textbf{0.9000} & 0.9985 & \textbf{0.9492} & \textbf{0.8128} & \textbf{0.7855} & \textbf{0.7832} \\
\midrule
\multirow{7}{*}{\textbf{DocSplit-Poly-Int}} 
& Claude Haiku 4.5   & 0.8709 & 0.9837 & 0.9273 & 0.7845 & 0.6707 & 0.6564 \\
& Claude Sonnet 4.5  & 0.8541 & \textbf{0.9942} & 0.9242 & 0.7808 & 0.6571 & 0.6525 \\
& DeepSeek V3.1      & 0.8752 & 0.9980 & 0.9366 & 0.8041 & 0.6700 & 0.6677 \\
& Gemma 3 27B        & 0.7596 & 0.9960 & 0.8778 & 0.5031 & 0.1911 & 0.1911 \\
& Qwen 3 VL 235B& \textbf{0.8893} & 0.9923 & \textbf{0.9408} & \textbf{0.8112} & \textbf{0.7349} & \textbf{0.7224} \\
\midrule
\multirow{7}{*}{\textbf{DocSplit-Poly-Rand}} 
& Claude Haiku 4.5   & 0.8351 & 0.9907 & 0.9129 & 0.7494 & 0.5457 & 0.5333 \\
& Claude Sonnet 4.5  & 0.8264 & 0.9962 & 0.9113 & 0.\textbf{7702} & 0.5703 & 0.5678 \\
& DeepSeek V3.1      & 0.8452 & 0.9888 & 0.9170 & 0.7593 & 0.5138 & 0.5021 \\
& Gemma 3 27B        & 0.7702 & \textbf{0.9980} & 0.8841 & 0.5106 & 0.1738 & 0.1738 \\
& Qwen 3 VL 235B& \textbf{0.8682} & 0.9793 & \textbf{0.9238} & 0.7672 & \textbf{0.6286} & \textbf{0.6022} \\
\bottomrule
\end{tabular}
\caption{Experimental results of document packet splitting task across all benchmarks. Clustering and Ordering are the components of the proposed Packet score, Page = Page level class accuracy, Page+Split = Page and split level accuracy, Page+Split+Order = Page and split level accuracy with order. Best results per benchmark shown in \textbf{bold}. For the Proposed metrics $\alpha$ = $\beta$ = $w$ = 0.5 were used.}
\label{tab:all_benchmarks}
\end{table*}


Table~\ref{tab:all_benchmarks} presents the evaluation of five vision-language models across five benchmark configurations, demonstrating how general-purpose models can address document packet splitting without task-specific training. Qwen 3 consistently achieves the highest Packet scores (0.9238 -- 0.9492), with clustering exceeding 0.86 and good ordering score (>0.97) compare to others. Claude Sonnet, Claude Haiku, and DeepSeek form a competitive second tier (Packet >0.90). Gemma 3 exhibits the weakest performance, with clustering as low as 0.5584; despite high ordering scores (>0.99), its poor clustering reveals boundary detection limitations.

\textbf{Impact of Benchmark Complexity}: The benchmarks represent increasing difficulty. \textit{DocSplit-Mono-Seq} yields the highest scores (Packet >0.93) due to document homogeneity and page continuity. \textit{DocSplit-Mono-Rand} introduces random ordering; models maintain comparable clustering (1--3\% degradation) while ordering remains high (>0.99), suggesting effective sequence recovery via content cues. Multi-category benchmarks show improved clustering from diverse document types providing more discriminative boundary features, but this advantage diminishes under page manipulation. \textit{DocSplit-Poly-Int} and \textit{DocSplit-Poly-Rand} present the most challenging conditions, with classical metrics degrading 20--30\% for weaker models.

\textbf{Clustering vs. Ordering Performance}: Ordering scores remain high across models (>0.97), indicating reliable sequence determination via structural cues once clustering succeeds. Clustering exhibits greater variance (0.56--0.90) and is the primary differentiator, as boundary detection requires holistic understanding of document semantics. This decoupling validates our framework and identifies boundary detection as the key challenge.

\textbf{Proposed vs. Classical Metrics}: Classical metrics exhibit fundamental 
limitations: (1) \textit{binary evaluation} with no partial credit, where 
DeepSeek achieves Clustering of 0.85 but Page+Split drops to 0.51 on 
\textit{DocSplit-Poly-Rand}, excessively penalizing near-correct predictions; 
(2) \textit{cascading penalty effect}, as a single boundary error propagates 
incorrect group IDs to all subsequent pages; (3) \textit{poor utility signals}: 
Gemma 3 on \textit{DocSplit-Poly-Rand} achieves Clustering of 0.77 yet 
Page+Split of only 0.17, suggesting near-complete failure despite reasonable 
grouping quality; (4) \textit{metric redundancy}, where Gemma 3 shows identical 
Page+Split and Page+Split+Order values across all benchmarks (e.g., 0.31 on 
\textit{DocSplit-Mono-Seq}), confirming ordering adds no discriminative value.

Our proposed metrics address these limitations through partial correctness quantification and task decomposition. Separate Clustering and Ordering scores provide diagnostic insights, while the unified Packet score better reflects practical utility by appropriately crediting near-correct predictions.

\section{Conclusion}

We introduce \textit{DocSplit}, the first comprehensive benchmark and novel evaluation framework for document packet splitting, an impactful yet underexplored task in intelligent document processing that addresses real-world challenges in healthcare claims, mortgage processing, and legal document management where heterogeneous multi-page packets must be decomposed into constituent documents. Our contributions are threefold: first, we formalize the \textit{DocSplit} task requiring simultaneous document boundary detection, type classification, and page sequence reconstruction from potentially shuffled and interleaved packets; second, we release five systematically designed benchmark datasets that progressively introduce complexity factors spanning the practical difficulty spectrum; and third, we propose a novel evaluation framework combining clustering metrics with ordering metrics into a unified Packet score that provides continuous, granular assessment superior to traditional binary exact matching approaches. Experiments across seven vision language models reveal that document boundary detection remains the primary bottleneck (0.56 to 0.90 clustering scores), identifying a clear direction for future research in cross-page consistency modeling and hierarchical document representations. We release all datasets, evaluation code, and experimental configurations to enable the research community to systematically advance document packet splitting methodologies.
\section*{Ethical Considerations}

\textbf{Dataset Composition and Privacy}: Our benchmark datasets are derived from the publicly available RVL-CDIP collection, which predominantly contains English-language documents from North American business and governmental sources. While this reflects common document processing scenarios, we acknowledge potential limitations in generalizability to low resources language documents and non-Western formats. All source documents undergo systematic preprocessing to remove potentially sensitive information, including personal identifiers, financial details, and confidential business information. Ground truth annotations are generated through automated processes based on original document metadata, with human verification conducted under fair compensation and clear task guidelines. We release all datasets, evaluation code, and experimental configurations to ensure transparency and reproducibility, enabling community-driven improvements to document packet splitting methodologies.

\textbf{Intended Use and Responsible Deployment}: {\ds} is designed to advance document understanding capabilities for legitimate applications in legal, financial, healthcare, and administrative domains where accurate document organization is critical. We recognize that document packet splitting technologies could potentially be misused for unauthorized document access or manipulation. We strongly encourage responsible deployment practices, including appropriate access controls, audit logging, and human oversight in high-stakes applications. Our evaluation framework is model-agnostic and designed to assess capabilities rather than promote specific commercial systems, supporting equitable comparison across diverse modeling approaches. Researchers and practitioners should implement safeguards commensurate with their specific use cases and regulatory requirements to ensure ethical application of these technologies.

\bibliography{anthology,custom}

\begin{thebibliography}{42}
\expandafter\ifx\csname natexlab\endcsname\relax\def\natexlab#1{#1}\fi

\bibitem[{Biten et~al.(2022)Biten, Tito, Gomez, Valveny, and Karatzas}]{biten2022ocr}
Ali~Furkan Biten, Ruben Tito, Lluis Gomez, Ernest Valveny, and Dimosthenis Karatzas. 2022.
\newblock Ocr-idl: Ocr annotations for industry document library dataset.
\newblock \emph{arXiv preprint arXiv:2202.12985}.

\bibitem[{Caffagni et~al.(2025)Caffagni, Sarto, Cornia, Baraldi, and Cucchiara}]{Caffagni2025}
Davide Caffagni, Sara Sarto, Marcella Cornia, Lorenzo Baraldi, and Rita Cucchiara. 2025.
\newblock Recurrence-enhanced vision-and-language transformers for robust multimodal document retrieval.
\newblock In \emph{Proceedings of the IEEE/CVF Conference on Computer Vision and Pattern Recognition (CVPR)}, pages 6262--6272.

\bibitem[{Chen et~al.(2025)Chen, Xu, Fei, Feng, and Elhoseiny}]{Chen2025}
Jun Chen, Dannong Xu, Junjie Fei, Chun-Mei Feng, and Mohamed Elhoseiny. 2025.
\newblock Document haystacks: Vision-language reasoning over piles of 1000+ documents.
\newblock In \emph{Proceedings of the IEEE/CVF Conference on Computer Vision and Pattern Recognition (CVPR)}, pages 24817--24826.

\bibitem[{Dimmick et~al.(1992)Dimmick, Garris, and Wilson}]{dimmick1992nist}
DL~Dimmick, MD~Garris, and CL~Wilson. 1992.
\newblock Nist special database 6. structured forms database 2.
\newblock Technical report, Technical report, National Institute of Standards and Technology.

\bibitem[{Duan et~al.(2025)Duan, Chen, Hu, Wang, Ye, Shi, Lu, Hou, Lu, Li, Dai, and Wang}]{Duan2025}
Yuchen Duan, Zhe Chen, Yusong Hu, Weiyun Wang, Shenglong Ye, Botian Shi, Lewei Lu, Qibin Hou, Tong Lu, Hongsheng Li, Jifeng Dai, and Wenhai Wang. 2025.
\newblock Docopilot: Improving multimodal models for document-level understanding.
\newblock In \emph{Proceedings of the IEEE/CVF Conference on Computer Vision and Pattern Recognition (CVPR)}, pages 4026--4037.

\bibitem[{Guo et~al.(2025)Guo, Qin, Yang, Zhang, Zeng, Li, and Lin}]{Guo2025}
Hao Guo, Xugong Qin, Jun Jie~Ou Yang, Peng Zhang, Gangyan Zeng, Yubo Li, and Hailun Lin. 2025.
\newblock Towards natural language-based document image retrieval: New dataset and benchmark.
\newblock In \emph{Proceedings of the IEEE/CVF Conference on Computer Vision and Pattern Recognition (CVPR)}, pages 29722--29732.

\bibitem[{Harley et~al.(2015{\natexlab{a}})Harley, Ufkes, and Derpanis}]{harley2015evaluation}
Adam~W Harley, Alex Ufkes, and Konstantinos~G Derpanis. 2015{\natexlab{a}}.
\newblock Evaluation of deep convolutional nets for document image classification and retrieval.
\newblock In \emph{2015 13th International Conference on Document Analysis and Recognition (ICDAR)}, pages 991--995. IEEE.

\bibitem[{Harley et~al.(2015{\natexlab{b}})Harley, Ufkes, and Derpanis}]{harley2015icdar}
Adam~W Harley, Alex Ufkes, and Konstantinos~G Derpanis. 2015{\natexlab{b}}.
\newblock \href {https://adamharley.com/icdar15/} {Evaluation of deep convolutional nets for document image classification and retrieval}.
\newblock In \emph{International Conference on Document Analysis and Recognition ({ICDAR})}.

\bibitem[{Hubert and Arabie(1985)}]{hubert1985comparing}
Lawrence Hubert and Phipps Arabie. 1985.
\newblock Comparing partitions.
\newblock \emph{Journal of classification}, 2(1):193--218.

\bibitem[{Jaume et~al.(2019)Jaume, Ekenel, and Thiran}]{jaume2019funsd}
Guillaume Jaume, Hazim~Kemal Ekenel, and Jean-Philippe Thiran. 2019.
\newblock Funsd: A dataset for form understanding in noisy scanned documents.
\newblock In \emph{2019 International Conference on Document Analysis and Recognition Workshops (ICDARW)}, volume~2, pages 1--6. IEEE.

\bibitem[{Jiang et~al.(2025)Jiang, Lin, Li, Liu, Yeh, and Chen}]{Jiang2025}
Zi-Han Jiang, Chien-Wei Lin, Wei-Hua Li, Hsuan-Tung Liu, Yi-Ren Yeh, and Chu-Song Chen. 2025.
\newblock Relation-rich visual document generator for visual information extraction.
\newblock In \emph{Proceedings of the IEEE/CVF Conference on Computer Vision and Pattern Recognition (CVPR)}.

\bibitem[{Kendall(1938)}]{kendall1938new}
Maurice~G Kendall. 1938.
\newblock A new measure of rank correlation.
\newblock \emph{Biometrika}, 30(1-2):81--93.

\bibitem[{Kumar and Doermann(2013)}]{kumar2013unsupervised}
Jayant Kumar and David Doermann. 2013.
\newblock Unsupervised classification of structurally similar document images.
\newblock In \emph{2013 12th International Conference on Document Analysis and Recognition}, pages 1225--1229. IEEE.

\bibitem[{Landeghem et~al.(2024)Landeghem, Biswas, Blaschko, and Moens}]{VanLandeghem2024}
Jordy~Van Landeghem, Sanket Biswas, Matthew~B. Blaschko, and Marie-Francine Moens. 2024.
\newblock Beyond document page classification: Design, datasets, and challenges.
\newblock In \emph{Proceedings of the IEEE/CVF Winter Conference on Applications of Computer Vision (WACV)}, pages 2950--2960.

\bibitem[{Larson et~al.(2022)Larson, Lim, Ai, Kuang, and Leach}]{larson2022evaluating}
Stefan Larson, Gordon Lim, Yutong Ai, David Kuang, and Kevin Leach. 2022.
\newblock Evaluating out-of-distribution performance on document image classifiers.
\newblock In \emph{Thirty-sixth Conference on Neural Information Processing Systems Datasets and Benchmarks Track}.

\bibitem[{Larson et~al.(2023)Larson, Lim, and Leach}]{larson2023evaluationdocumentclassificationusing}
Stefan Larson, Gordon Lim, and Kevin Leach. 2023.
\newblock \href {http://arxiv.org/abs/2306.12550} {On evaluation of document classification using rvl-cdip}.

\bibitem[{Lewis et~al.(2006)Lewis, Agam, Argamon, Frieder, Grossman, and Heard}]{lewis2006building}
David Lewis, Gady Agam, Shlomo Argamon, Ophir Frieder, David Grossman, and Jefferson Heard. 2006.
\newblock Building a test collection for complex document information processing.
\newblock In \emph{Proceedings of the 29th annual international ACM SIGIR conference on Research and development in information retrieval}, pages 665--666.

\bibitem[{Li et~al.(2020)Li, Xu, Cui, Huang, Wei, Li, and Zhou}]{li2020docbank}
Minghao Li, Yiheng Xu, Lei Cui, Shaohan Huang, Furu Wei, Zhoujun Li, and Ming Zhou. 2020.
\newblock \href {http://arxiv.org/abs/2006.01038} {Docbank: A benchmark dataset for document layout analysis}.

\bibitem[{Li et~al.(2025)Li, Yin, and Liu}]{li2025docsam}
Xinhao Li, Fei Yin, and Cheng-Lin Liu. 2025.
\newblock Docsam: Unified document image segmentation via query decomposition and heterogeneous mixed learning.
\newblock In \emph{Proceedings of the IEEE/CVF Conference on Computer Vision and Pattern Recognition (CVPR)}, pages 4026--4037.

\bibitem[{Liao et~al.(2024)Liao, Wang, Li et~al.}]{Liao2024doclayllm}
Wenhui Liao, Jiapeng Wang, Hongliang Li, et~al. 2024.
\newblock Doclayllm: An efficient and effective multi-modal extension of large language models for text-rich document understanding.
\newblock \emph{arXiv preprint arXiv:2408.15045}.

\bibitem[{Lior et~al.(2024)Lior, Caciularu, Cattan, Levy, Shapira, and Dagan}]{lior2024seam}
Guy Lior, Asaf Caciularu, Ariel Cattan, Shai Levy, Omer Shapira, and Ido Dagan. 2024.
\newblock Seam: A stochastic benchmark for multi-document tasks.
\newblock \emph{arXiv preprint arXiv:2406.16086}.

\bibitem[{Long et~al.(2005)Long, Antani, and Thoma}]{long2005image}
L~Rodney Long, Sameer~K Antani, and George~R Thoma. 2005.
\newblock Image informatics at a national research center.
\newblock \emph{Computerized Medical Imaging and Graphics}, 29(2-3):171--193.

\bibitem[{Luo et~al.(2024)Luo, Shen, Zhu, Zheng, Yu, and Yao}]{Luo2024}
Chuwei Luo, Yufan Shen, Zhaoqing Zhu, Qi~Zheng, Zhi Yu, and Cong Yao. 2024.
\newblock Layoutllm: Layout instruction tuning with large language models for document understanding.
\newblock In \emph{Proceedings of the IEEE/CVF Conference on Computer Vision and Pattern Recognition (CVPR)}, pages 15630--15640.

\bibitem[{Luo et~al.(2025)Luo, Tang, Zheng, Yao, Jin, Li, Xue, and Si}]{luo2025bivldocbidirectionalvisionlanguagemodeling}
Chuwei Luo, Guozhi Tang, Qi~Zheng, Cong Yao, Lianwen Jin, Chenliang Li, Yang Xue, and Luo Si. 2025.
\newblock \href {http://arxiv.org/abs/2206.13155} {Bi-vldoc: Bidirectional vision-language modeling for visually-rich document understanding}.

\bibitem[{Mathew et~al.(2020)Mathew, Tito, Karatzas, Manmatha, and Jawahar}]{mathew2020document}
Minesh Mathew, Ruben Tito, Dimosthenis Karatzas, R~Manmatha, and CV~Jawahar. 2020.
\newblock Document visual question answering challenge 2020.
\newblock \emph{arXiv preprint arXiv:2008.08899}.

\bibitem[{Mungmeeprued et~al.(2022)Mungmeeprued, Ma, Mehta, and Lipani}]{mungmeeprued2022tab}
Thisanaporn Mungmeeprued, Yuxin Ma, Nisarg Mehta, and Aldo Lipani. 2022.
\newblock Tab this folder of documents: page stream segmentation of business documents.
\newblock In \emph{Proceedings of the 22nd ACM Symposium on Document Engineering}, pages 1--10.

\bibitem[{Ouyang et~al.(2025)Ouyang, Qu, Zhou, Zhu, Zhang, Lin, Wang, Zhao, Jiang, Zhao, Shi, Wu, Chu, Liu, Li, Xu, Zhang, Shi, Tu, and He}]{Ouyang2025}
Linke Ouyang, Yuan Qu, Hongbin Zhou, Jiawei Zhu, Rui Zhang, Qunshu Lin, Bin Wang, Zhiyuan Zhao, Man Jiang, Xiaomeng Zhao, Jin Shi, Fan Wu, Pei Chu, Minghao Liu, Zhenxiang Li, Chao Xu, Bo~Zhang, Botian Shi, Zhongying Tu, and Conghui He. 2025.
\newblock Omnidocbench: Benchmarking diverse pdf document parsing with comprehensive annotations.
\newblock In \emph{Proceedings of the IEEE/CVF Conference on Computer Vision and Pattern Recognition (CVPR)}, pages 24838--24848.

\bibitem[{{\v{S}}imsa et~al.(2023){\v{S}}imsa, {\v{S}}ulc, U{\v{r}}i{\v{c}}{\'a}{\v{r}}, Patel, Hamdi, Koci{\'a}n, Skalick{\`y}, Matas, Doucet, Coustaty et~al.}]{simsa2023docile}
{\v{S}}t{\v{e}}p{\'a}n {\v{S}}imsa, Milan {\v{S}}ulc, Michal U{\v{r}}i{\v{c}}{\'a}{\v{r}}, Yash Patel, Ahmed Hamdi, Mat{\v{e}}j Koci{\'a}n, Maty{\'a}{\v{s}} Skalick{\`y}, Ji{\v{r}}{\'\i} Matas, Antoine Doucet, Micka{\"e}l Coustaty, et~al. 2023.
\newblock Docile benchmark for document information localization and extraction.
\newblock \emph{arXiv preprint arXiv:2302.05658}.

\bibitem[{Stanislawek et~al.(2021)Stanislawek, Gralinski, Wr{\'{o}}blewska, Lipinski, Kaliska, Rosalska, Topolski, and Biecek}]{kleisterStanislawekGWLK21}
Tomasz Stanislawek, Filip Gralinski, Anna Wr{\'{o}}blewska, Dawid Lipinski, Agnieszka Kaliska, Paulina Rosalska, Bartosz Topolski, and Przemyslaw Biecek. 2021.
\newblock \href {https://doi.org/10.1007/978-3-030-86549-8\_36} {Kleister: Key information extraction datasets involving long documents with complex layouts}.
\newblock In \emph{ICDAR}, volume 12821 of \emph{Lecture Notes in Computer Science}, pages 564--579. Springer.

\bibitem[{Stray and Svetlichnaya()}]{straydeepform}
J~Stray and S~Svetlichnaya.
\newblock Deepform: extract information from documents (2020).

\bibitem[{Turski et~al.(2023)Turski, Stanis{\l}awek, Kaczmarek, Dyda, and Grali{\'n}ski}]{turski2023ccpdf}
Micha{\l} Turski, Tomasz Stanis{\l}awek, Karol Kaczmarek, Pawe{\l} Dyda, and Filip Grali{\'n}ski. 2023.
\newblock Ccpdf: Building a high quality corpus for visually rich documents from web crawl data.
\newblock \emph{arXiv preprint arXiv:2304.14953}.

\bibitem[{Van~Landeghem et~al.(2024)Van~Landeghem, Biswas, Blaschko, and Moens}]{van2024beyond}
Jordy Van~Landeghem, Sanket Biswas, Matthew Blaschko, and Marie-Francine Moens. 2024.
\newblock Beyond document page classification: design, datasets, and challenges.
\newblock In \emph{Proceedings of the IEEE/CVF Winter Conference on Applications of Computer Vision}, pages 2962--2972.

\bibitem[{Van~Landeghem et~al.(2023)Van~Landeghem, Tito, Borchmann, Pietruszka, Joziak, Powalski, Jurkiewicz, Coustaty, Ackaert, Valveny, Blaschko, Moens, and Stanislawek}]{vanlandeghem2023document}
Jordy Van~Landeghem, Rub\`{e}n Tito, {\L}ukasz Borchmann, Micha{\l} Pietruszka, Pawel Joziak, Rafal Powalski, Dawid Jurkiewicz, Mickael Coustaty, Bertrand Ackaert, Ernest Valveny, Matthew~B. Blaschko, Marie-Francine Moens, and Tomasz Stanislawek. 2023.
\newblock {Document Understanding Dataset and Evaluation (DUDE)}.
\newblock In \emph{International Conference on Computer Vision}.

\bibitem[{Wang et~al.(2025{\natexlab{a}})Wang, Gao, Xiao, Huang, Si, and Hughes}]{wang2025document}
Zihan Wang, Chao Gao, Cheng Xiao, Yun Huang, Si~Si, and Macduff Hughes. 2025{\natexlab{a}}.
\newblock Document segmentation matters for retrieval-augmented generation.
\newblock In \emph{Findings of the Association for Computational Linguistics: ACL 2025}.

\bibitem[{Wang et~al.(2025{\natexlab{b}})Wang, Guan, Fu, Duan, Jiang, Guo, Guo, Luo, Shen, and Yang}]{Wang2025marten}
Zining Wang, Tongkun Guan, Pei Fu, Chen Duan, Qianyi Jiang, Zhentao Guo, Shan Guo, Junfeng Luo, Wei Shen, and Xiaokang Yang. 2025{\natexlab{b}}.
\newblock Marten: Visual question answering with mask generation for multi-modal document understanding.
\newblock \emph{arXiv preprint arXiv:2503.14140}.

\bibitem[{Zhang et~al.(2024)Zhang, Yang, Lai, Xing, Feng, and Sun}]{Zhang2024dockylin}
Jiaxin Zhang, Wentao Yang, Songxuan Lai, Yan Xing, Feiyue Feng, and Yipeng Sun. 2024.
\newblock Dockylin: A large multimodal model for visual document understanding with efficient visual slimming.
\newblock \emph{arXiv preprint arXiv:2406.19101}.

\bibitem[{Zheng et~al.(2021)Zheng, Burdick, Popa, Zhong, and Wang}]{zheng2021global}
Xinyi Zheng, Douglas Burdick, Lucian Popa, Xu~Zhong, and Nancy Xin~Ru Wang. 2021.
\newblock Global table extractor (gte): A framework for joint table identification and cell structure recognition using visual context.
\newblock In \emph{Proceedings of the IEEE/CVF winter conference on applications of computer vision}, pages 697--706.

\bibitem[{Zhong et~al.(2020)Zhong, ShafieiBavani, and Jimeno~Yepes}]{zhong2020image}
Xu~Zhong, Elaheh ShafieiBavani, and Antonio Jimeno~Yepes. 2020.
\newblock Image-based table recognition: data, model, and evaluation.
\newblock In \emph{Computer Vision--ECCV 2020: 16th European Conference, Glasgow, UK, August 23--28, 2020, Proceedings, Part XXI 16}, pages 564--580. Springer.

\bibitem[{Zhong et~al.(2019)Zhong, Tang, and Yepes}]{zhong2019publaynet}
Xu~Zhong, Jianbin Tang, and Antonio~Jimeno Yepes. 2019.
\newblock Publaynet: largest dataset ever for document layout analysis.
\newblock In \emph{2019 International Conference on Document Analysis and Recognition (ICDAR)}, pages 1015--1022. IEEE.

\bibitem[{Zhu et~al.(2022)Zhu, Lei, Feng, Wang, Zhang, and Chua}]{Zhu_2022}
Fengbin Zhu, Wenqiang Lei, Fuli Feng, Chao Wang, Haozhou Zhang, and Tat-Seng Chua. 2022.
\newblock \href {https://doi.org/10.1145/3503161.3548422} {Towards complex document understanding by discrete reasoning}.
\newblock In \emph{Proceedings of the 30th {ACM} International Conference on Multimedia}. {ACM}.

\bibitem[{Zhu and Doermann(2007)}]{zhu2007automatic}
Guangyu Zhu and David Doermann. 2007.
\newblock Automatic document logo detection.
\newblock In \emph{Ninth International Conference on Document Analysis and Recognition (ICDAR 2007)}, volume~2, pages 864--868. IEEE.

\bibitem[{Zhu et~al.(2025)Zhu, Luo, Shao, Gao, Xing, Zheng, and Zhang}]{Zhu2025layouttoken}
Zhaoqing Zhu, Chuwei Luo, Zirui Shao, Feiyu Gao, Hangdi Xing, Qi~Zheng, and Ji~Zhang. 2025.
\newblock A simple yet effective layout token in large language models for document understanding.
\newblock \emph{arXiv preprint arXiv:2503.18434}.

\end{thebibliography}
\bibliographystyle{acl_natbib}


\onecolumn
\appendix






\section{Robustness of Proposed Evaluation Metric: Edge Cases Evaluation}
\label{appendix:edge_cases}

This appendix evaluates the proposed document packet metrics against classical metrics using 10 carefully designed edge cases (Table~\ref{tab:edge_cases_description}). Each test case contains 5 pages (3 ``invoice'' pages and 2 ``form'' pages) to isolate specific error types and their impact on evaluation scores. Tables~\ref{tab:proposed_results} and~\ref{tab:classical_results} present the results for proposed and classical metrics respectively, with a direct comparison in Table~\ref{tab:comparison}.

\begin{table}[htbp]
\centering
\small
\begin{tabular}{@{}lp{7.5cm}@{}}
\toprule
\textbf{Test Case} & \textbf{Description} \\
\midrule
Perfect & All predictions match ground truth \\
Misclassification Only & Correct grouping and ordering, but class labels swapped \\
Wrong Grouping Only & Correct classification and ordering, but group IDs swapped \\
Wrong Ordering Only & Correct classification and grouping, but page order scrambled \\
Split Groups & One document split into two groups (over-segmentation) \\
Merged Groups & Two documents merged into one group (under-segmentation) \\
Partial Misclass & One page misclassified within otherwise correct group \\
Multiple Errors & Combination of classification, grouping, and ordering errors \\
Duplicate Page Nums & Same page number predicted for multiple pages \\
Reverse Order & Pages in completely reversed sequence \\
\bottomrule
\end{tabular}
\caption{Edge case test scenarios. Each case isolates a specific error type: classification errors (Misclassification, Partial Misclass), boundary detection errors (Split Groups, Merged Groups, Wrong Grouping), ordering errors (Wrong Ordering, Reverse Order, Duplicate Page Nums), or combinations thereof (Multiple Errors). The Perfect case serves as the baseline with all predictions matching ground truth.}
\label{tab:edge_cases_description}
\end{table}

\begin{table}[htbp]
\centering
\small
\begin{tabular}{@{}lccccc@{}}
\toprule
\textbf{Test Case} & $\mathbf{S_{\text{packet}}}$ & $\mathbf{S_{\text{clustering}}}$ & \textbf{V-measure} & \textbf{RI} & $\mathbf{S_{\text{ordering}}}$ \\
\midrule
Perfect & 1.0000 & 1.0000 & 1.0000 & 1.0000 & 1.0000 \\
Misclassification Only & 0.7974 & 0.5949 & 0.5897 & 0.6000 & 1.0000 \\
Wrong Grouping Only & 1.0000 & 1.0000 & 1.0000 & 1.0000 & 1.0000 \\
Wrong Ordering Only & 0.1667 & 1.0000 & 1.0000 & 1.0000 & $-0.6667$ \\
Split Groups & 0.8428 & 0.6856 & 0.6713 & 0.7000 & 1.0000 \\
Merged Groups & 0.6000 & 0.2000 & 0.0000 & 0.4000 & 1.0000 \\
Partial Misclass & 0.8947 & 0.7895 & 0.7790 & 0.8000 & 1.0000 \\
Multiple Errors & $-0.0359$ & 0.5949 & 0.5897 & 0.6000 & $-0.6667$ \\
Duplicate Page Nums & 0.9541 & 1.0000 & 1.0000 & 1.0000 & 0.9082 \\
Reverse Order & 0.0000 & 1.0000 & 1.0000 & 1.0000 & $-1.0000$ \\
\bottomrule
\end{tabular}
\caption{Proposed metrics results ($\alpha = \beta = w = 0.5$). $S_{\text{packet}}$ is the combined score, $S_{\text{clustering}}$ combines V-measure and Rand Index (RI), and $S_{\text{ordering}}$ uses Kendall's Tau ($\tau$). Key observations: (1) classification errors reduce $S_{\text{clustering}}$ proportionally (Misclassification: 0.59), (2) ordering errors produce negative $S_{\text{ordering}}$ (Reverse Order: $-1.0$), (3) swapped group IDs receive full credit when structure is preserved (Wrong Grouping: 1.0), and (4) combined errors yield low or negative $S_{\text{packet}}$ (Multiple Errors: $-0.04$).}
\label{tab:proposed_results}
\end{table}

\begin{table}[htbp]
\centering
\small
\begin{tabular}{@{}lccc@{}}
\toprule
\textbf{Test Case} & \textbf{Page} & \textbf{Page+Split} & \textbf{Page+Split+Order} \\
\midrule
Perfect & 100.00\% & 100.00\% & 100.00\% \\
Misclassification Only & 0.00\% & 0.00\% & 0.00\% \\
Wrong Grouping Only & 100.00\% & 100.00\% & 100.00\% \\
Wrong Ordering Only & 100.00\% & 100.00\% & 0.00\% \\
Split Groups & 100.00\% & 0.00\% & 0.00\% \\
Merged Groups & 100.00\% & 0.00\% & 0.00\% \\
Partial Misclass & 80.00\% & 50.00\% & 50.00\% \\
Multiple Errors & 20.00\% & 0.00\% & 0.00\% \\
Duplicate Page Nums & 100.00\% & 50.00\% & 50.00\% \\
Reverse Order & 100.00\% & 100.00\% & 0.00\% \\
\bottomrule
\end{tabular}
\caption{Classical metrics results using exact matching at three strictness levels: Page (classification only), Page+Split (classification and boundary detection), and Page+Split+Order (all three aspects). Key limitations: (1) binary outcomes with no partial credit (Misclassification: 0\% despite correct grouping), (2) Split Groups and Merged Groups both score 0\% despite different error severities, and (3) Reverse Order scores identically to Wrong Ordering (0\%) despite being more severe.}
\label{tab:classical_results}
\end{table}

\begin{table}[htbp]
\centering
\small
\begin{tabular}{@{}lp{4.8cm}p{5.2cm}@{}}
\toprule
\textbf{Test Case} & \textbf{Classical} & \textbf{Proposed} \\
\midrule
Misclassification & All = 0\% (complete failure) & $S_{\text{clustering}}$ = 0.59, $S_{\text{packet}}$ = 0.80 \\
Split Groups & Page+Split = 0\% & $S_{\text{clustering}}$ = 0.69, $S_{\text{packet}}$ = 0.84 \\
Merged Groups & Page+Split = 0\% & $S_{\text{clustering}}$ = 0.20, V-measure = 0 \\
Reverse Order & Page+Split+Order = 0\% & $S_{\text{ordering}}$ = $-1.0$ (captures direction) \\
Partial Misclass & Page+Split = 50\% & $S_{\text{packet}}$ = 0.89 (localized penalty) \\
\bottomrule
\end{tabular}
\caption{Comparative analysis of classical vs.\ proposed metrics. The proposed framework provides: (1) continuous scores instead of binary pass/fail, (2) proportional penalties reflecting error severity (Split vs.\ Merged receive different scores), (3) directional ordering information ($S_{\text{ordering}}$ distinguishes scrambled from reversed), and (4) localized impact for partial errors.}
\label{tab:comparison}
\end{table}

\subsection*{Robustness Against Classical Failure Cases}

The edge cases reveal specific scenarios where classical metrics fail to provide meaningful evaluation, while the proposed metrics remain robust (see Tables~\ref{tab:proposed_results}--\ref{tab:comparison}):

\textbf{Partial Correctness.} Classical metrics report complete failure (0\% on all metrics) when grouping and ordering are correct but classification labels differ (Misclassification Only). The proposed metrics recognize preserved structure: $S_{\text{packet}}$ = 0.80, $S_{\text{clustering}}$ = 0.59.

\textbf{Error Severity Distinction.} Classical metrics score Split Groups and Merged Groups identically (0\% on Page+Split), despite different practical implications: under-segmentation loses document boundaries permanently while over-segmentation preserves them. Proposed metrics appropriately distinguish: $S_{\text{clustering}}$ = 0.69 (Split) vs.\ 0.20 (Merged).

\textbf{Ordering Direction.} Classical metrics collapse Reverse Order and Wrong Ordering to the same score (0\% on Page+Split+Order), failing to capture that complete reversal is more severe than partial scrambling. Proposed metrics distinguish via Kendall's Tau: $S_{\text{ordering}}$ = $-1.0$ (Reverse) vs.\ $-0.67$ (Wrong Ordering).

\textbf{Localized Error Impact.} Classical metrics cascade single-page errors across entire groups (Partial Misclass: Page+Split = 50\%). Proposed metrics localize the penalty: $S_{\text{packet}}$ = 0.89, reflecting that only one page was affected.

These robustness properties, including continuous scoring, severity-aware penalties, directional ordering information, and localized error impact, make the proposed framework more suitable for document packet splitting evaluation.

\section{Dataset Creation (Detailed Version)}
\label{appendix:dataset_curation}

This appendix provides extended descriptions of each benchmark dataset introduced in Section~\ref{sec:dataset_curation}.

\subsection{Data Source and Preprocessing}

Our benchmark datasets are derived from the RVL-CDIP-MP dataset, a multi-page extension of the original RVL-CDIP. We performed systematic data cleaning by removing corrupted files, yielding 995 valid documents comprising 5,231 pages across 13 categories: form, scientific publication, handwritten, resume, letter, language, specification, questionnaire, memo, news article, email, invoice, and budget. The distribution ranges from 183 documents (resume) to 7 documents (language), providing diverse representation of real-world document types encountered in industries ranging from legal and financial services to healthcare.

Document preprocessing involved generating visual representations by converting each document page to an image format using Python, while textual information was extracted using a hybrid OCR approach. AWS Textract with LAYOUT, TABLES, FORMS, and SIGNATURES features enabled was used for most document categories, while multilingual documents in the "language" category were processed using DeepSeek OCR (deepseek-ai/DeepSeek-OCR), while multilingual documents in the "language" category were processed using DeepSeek OCR (deepseek-ai/
DeepSeek-OCR), a specialized vision-language model running on GPU infrastructure with flash-attention optimization for enhanced multilingual text recognition. Both representations were stored in a standardized format supporting HuggingFace datasets compatibility. A document mapping CSV was generated containing metadata (type, name, size, pages, validation status) to facilitate downstream benchmark creation.

\subsection{Dataset Splitting Strategy}

To prevent data leakage, we performed category-wise stratified splitting before benchmark generation. Documents within each category were allocated as: 55\% training, 20\% validation, and 25\% testing. This ensures no document appears across multiple splits, maintaining evaluation integrity for the Multi-page Document Packet Classification and Splitting (DocSplit) task.

\subsection{Benchmark Dataset Variations}

We created five distinct benchmarks\footref{fn:dataset} targeting different document concatenation scenarios that reflect real-world packet splitting challenges. These benchmarks systematically evaluate models' abilities to identify document boundaries, classify document types, and maintain correct page ordering—the three critical components of the DocSplit task.

\textbf{{\dsMonoSeq} (Single Category Document Concatenation Sequentially):} Creates document packets by first randomly selecting a single document category (excluding the 'language' category due to insufficient samples) and determining a target page count. Similar to Multi Category Document Concatenation, it selects and concatenates documents only from the chosen category until reaching the target page count. This benchmark addresses the challenging scenario where document boundaries must be detected without category transitions as discriminative signals. Such scenarios frequently occur in legal document processing where multiple contracts or agreements of the same type are bundled together, requiring models to identify subtle boundary indicators such as signature pages, document headers, or semantic shifts within homogeneous content.

\textbf{{\dsMonoRand} (Single Category Document Pages Randomization):} Similar to Single Category Document Concatenation, we randomly select a category (excluding 'language') and sample multiple documents from it. However, instead of maintaining the original page sequence, this benchmark randomly shuffles all pages from the selected documents. This benchmark evaluates models' robustness to page-level disruptions common in manual document assembly, particularly in transportation logistics where drivers photograph pages at delivery sites in arbitrary order. The task requires simultaneous document boundary detection and page sequence reconstruction, testing models' ability to leverage both local page features and global document coherence.

\textbf{{\dsPolySeq} (Multi Category Documents Concatenation Sequentially):} Creates document packets by first determining a target page count (5-20 pages), then sequentially selecting documents from different categories without repetition. For each selected document, all of its pages are included while preserving the original page ordering, and this process continues until the target page count is reached. This benchmark simulates the most common real-world scenario where heterogeneous documents are assembled into packets, as observed in medical claims processing where prescription records, laboratory results, and insurance forms are concatenated. The varying document types test models' ability to detect inter-document boundaries based on content and structural transitions, a fundamental requirement for accurate packet splitting.

\textbf{{\dsPolyInt} (Multi Category Document Pages Interleaving):} Similar to Multi Category Document Concatenation, in this benchmark we randomly select multiple documents from different categories until target page count is reached. Instead of concatenating whole documents sequentially, here we interleave pages from these documents in a round-robin fashion. For example, if three documents are selected, the resulting packet would have page 1 from document A, then page 1 from document B, then page 1 from document C, followed by page 2 from document A, and so on, until all pages are exhausted. This benchmark simulates scenarios where documents from various sources are merged during batch processing, as occurs in mortgage processing where property deeds, tax records, and legal notices are combined into a single packet with pages from different documents interspersed throughout. Unlike complete shuffling, this benchmark preserves the sequential ordering of pages within each document while mixing pages from different documents, challenging models to identify which non-contiguous pages belong together and reconstruct the original documents from the interleaved sequence.

\textbf{{\dsPolyRand} (Multi Category Document Pages Randomization):} This benchmark begins by selecting documents from various categories, similar to Multi Category Document Concatenation, then applies complete randomization across all pages from different categories, representing maximum entropy scenarios. This benchmark stress-tests model robustness under worst-case conditions where no structural assumptions hold, mirroring real-world failures in document management systems or emergency document recovery scenarios. The complete absence of ordering patterns requires models to rely solely on content-based features for both clustering pages into documents and reconstructing their original sequence, providing an upper bound on task difficulty.

\subsection{Ground Truth Annotation}
\label{app:annotation}

Each benchmark sample includes structured ground truth annotations essential for evaluating packet splitting performance:
\begin{itemize}
    \item \texttt{doc\_type}: Document category for classification evaluation
    \item \texttt{original\_doc\_name}: Original name of the source document
    \item \texttt{parent\_doc\_name}: Name of the parent document packet denoting single sample
    \item \texttt{local\_doc\_id}: Unique identifier for each document within a packet
    \item \texttt{page}: Page number within the concatenated document
    \item \texttt{image\_path}: Path to the page's image file
    \item \texttt{text\_path}: Path to the page's extracted text file
    \item \texttt{group\_id}: Identifier linking pages from the same source document
    \item \texttt{local\_doc\_id\_page\_ordinal}: Original page number within source document
\end{itemize}

This annotation schema enables comprehensive evaluation using our proposed metrics that combine clustering performance (Rand Index and V-measure) with ordering accuracy (Kendall's Tau), capturing both document boundary detection and page sequence reconstruction—the dual objectives of the MDCS task. All benchmarks utilize consistent JSON output format with unique categories allocated per split to ensure robust generalization assessment. 


\section{Dataset Statistics and Comparisons}

Table~\ref{tab:sources} provides a comprehensive comparison of DocSplit against existing document understanding benchmarks. The comparison encompasses datasets spanning multiple domains including industry documents, scientific publications, financial reports, and legal contracts. Existing benchmarks primarily address isolated tasks such as document classification (DC), document layout analysis (DLA), key information extraction (KIE), question answering (QA), and table structure recognition (TSR). In contrast, DocSplit uniquely addresses document packet splitting (DPS) alongside classification and key information extraction, reflecting the compound nature of real-world document processing pipelines. Notably, DocSplit provides both OCR annotations and layout information across all variants, enabling multimodal evaluation approaches. The dataset scale of 1.55M pages in the combined DocSplit benchmark positions it among the largest document understanding resources available, while the diversity of source materials from UCSF and Document Cloud ensures representation of authentic document variations encountered in enterprise settings.

Table~\ref{tab:packet-stats} presents detailed statistics characterizing the structural properties of each DocSplit variant. The benchmark systematically varies packet complexity through two dimensions: packet size (Large versus Small) and document arrangement strategy (sequential, interleaved, or randomized). Large variants contain substantially more pages per packet, ranging from 45 to 127 pages on average, compared to 11 to 15 pages in Small variants. This size differential enables evaluation of model scalability and context window utilization. The number of subdocuments per packet varies considerably, with Large variants containing 5 to 16 subdocuments while Small variants contain 3 to 4 subdocuments. Poly variants (PolySeq, PolyInt, PolyRand) incorporate multiple document types within each packet, averaging 7 to 8 distinct types in Large configurations and approximately 3.5 types in Small configurations. Mono variants (MonoSeq, MonoRand) maintain exactly one document type per packet, isolating the boundary detection challenge from cross-type discrimination. The pages per subdocument metric reveals that documents in Mono variants tend to be longer (8 to 16 pages) compared to Poly variants (5 to 8 pages), reflecting the different sampling strategies employed during benchmark construction.

Table~\ref{tab:doctype-dist} details the distribution of document types across all benchmark variants and data splits. The 13 document categories span diverse formats including structured documents (Form, Invoice, Budget), correspondence (Email, Letter, Memo), technical materials (Scientific Publication, Specification), and specialized formats (Questionnaire, Resume, Handwritten, News Article, Language). The Poly variants exhibit balanced representation across all categories, with each type contributing between 100 and 350 instances in test splits for Large variants. This balance ensures that evaluation metrics are not dominated by any single document type. The Mono variants deliberately exclude the Language category due to insufficient samples in the source corpus, as indicated by zero or near-zero counts for this type. Within Mono variants, document types are distributed according to their natural frequency in the source data, with Form, Letter, and Resume categories being most prevalent. The stratified splitting strategy ensures proportional representation across training, validation, and test partitions, preventing data leakage while maintaining statistical consistency for reliable model comparison.

\begin{table*}[!t]
\small
\centering
\begin{tabular}{@{}lllllcc@{}}

\toprule
Dataset      & Size & Data Source                 & Domain              & Task     & OCR & Layout \\ \midrule
IIT-CDIP~\cite{lewis2006building}     &  35.5M    & UCSF-IDL                    & Industry        & P       &  \xmark   & \xmark       \\
RVL-CDIP~\cite{harley2015evaluation}     &  400K    & UCSF-IDL                    & Industry        & DC     & \xmark    & \xmark     \\
RVL-CDIP-N~\cite{larson2022evaluating}     &  1K    & Doc Cloud                    & Industry        & DC     & \xmark    & \xmark     \\
TAB~\cite{mungmeeprued2022tab}     &  44.8K    & UCSF-IDL                    & Industry        & DC     & \xmark    & \xmark     \\
FUNSD~\cite{jaume2019funsd}        &  199    & UCSF-IDL                    & Industry        & KIE      & \cmark    & \xmark       \\
SP-DocVQA~\cite{mathew2020document}       &  12K    & UCSF-IDL                    & Industry        & QA       & \cmark    & \xmark       \\
OCR-IDL~\cite{biten2022ocr}      &  26M    & UCSF-IDL                    & Industry        & P &  \cmark   & \xmark       \\
FinTabNet~\cite{zheng2021global}    &  89.7K    & Annual Reports S\&P         & Finance             & TSR      &  \xmark   &   \cmark     \\
Kleister-NDA~\cite{kleisterStanislawekGWLK21} & 3.2K     & EDGAR                       & US NDAs             & KIE         & \cmark    & \xmark       \\
Kleister-Charity~\cite{kleisterStanislawekGWLK21} & 61.6K     &  UK Charity      & Legal             &   KIE       & \cmark    & \xmark       \\
DeepForm~\cite{straydeepform}    &  20K    & FCC Inspection & Forms     &  KIE        & \cmark    & \xmark       \\
TAT-QA~\cite{Zhu_2022}       &   2.8K   & Open WorldBank                            & Finance             & QA       & \cmark    &  \xmark      \\
PubLayNet~\cite{zhong2019publaynet}   & 360K     & PubMed Central              & Scientific  & DLA      & \xmark    & \cmark       \\
DocBank~\cite{li2020docbank}      & 500K     & arxiv                       & Scientific  & DLA         & \cmark    &  \cmark      \\
PubTabNet~\cite{zhong2020image}    &  568K    & PubMed Central              & Scientific  & TSR      & \xmark    & \cmark       \\
DUDE~\cite{vanlandeghem2023document}         &  40K    & Mixed             &  Multi                  &       QA   & \cmark    &  \xmark      \\
Docile~\cite{simsa2023docile}      &   106K   &     EDGAR \& synthetic                        &           Industry          &       KIE   & \cmark    & \xmark       \\ 
CC-PDF~\cite{turski2023ccpdf}         & 1.1M   & CC          &  Multi                &       P   & \xmark    &  \xmark      \\
\bottomrule
\textbf{\dsMonoSeq}-Small       & 91K   & UCSF \& Doc Cloud      &  Industry                  &       DPS, P, DC, KIE   & \cmark    &  \cmark      \\
\textbf{\dsMonoSeq}-Large       & 63K   & UCSF \& Doc Cloud      &  Industry                  &       DPS, P, DC, KIE   & \cmark    &  \cmark      \\
\textbf{\dsMonoRand}-Small       & 81K   & UCSF \& Doc Cloud      &  Industry                  &       DPS, P, DC, KIE   & \cmark    &  \cmark      \\
\textbf{\dsMonoRand}-Large       & 42K   & UCSF \& Doc Cloud      &  Industry                  &       DPS, P, DC, KIE   & \cmark    &  \cmark      \\
\textbf{\dsPolySeq}-Small       & 152K   & UCSF \& Doc Cloud      &  Industry                  &       DPS, P, DC, KIE   & \cmark    &  \cmark      \\
\textbf{\dsPolySeq}-Large       & 288K   & UCSF \& Doc Cloud      &  Industry                  &       DPS, P, DC, KIE   & \cmark    &  \cmark      \\
\textbf{\dsPolyInt}-Small       & 149K   & UCSF \& Doc Cloud      &  Industry                  &       DPS, P, DC, KIE   & \cmark    &  \cmark      \\
\textbf{\dsPolyInt}-Large       & 270K   & UCSF \& Doc Cloud      &  Industry                  &       DPS, P, DC, KIE   & \cmark    &  \cmark      \\
\textbf{\dsPolyRand}-Small       & 149K   & UCSF \& Doc Cloud      &  Industry                  &       DPS, P, DC, KIE   & \cmark    &  \cmark      \\
\textbf{\dsPolyRand}-Large       & 270K   & UCSF \& Doc Cloud      &  Industry                  &       DPS, P, DC, KIE   & \cmark    &  \cmark      \\
\hline
\textbf{\ds}-Combined       & 1.55M   & UCSF \& Doc Cloud      &  Industry                  &       DPS, P, DC, KIE   & \cmark    &  \cmark      \\
\bottomrule
\end{tabular}
\caption{Document understanding benchmarks with corresponding data sources, domains, and task specifications. DPS: Document Packet Splitting, DC: Document Classification, DLA: Document Layout Analysis, KIE: Key Information Extraction, QA: Question Answering, TSR: Table Structure Recognition, P: Pretrain}
\label{tab:sources}
\end{table*}
\begin{table}[!t]
\centering
\caption{Packet statistics across DocSplit benchmark variants. Large variants contain 45 to 127 pages per packet with 5 to 16 subdocuments, while Small variants contain 11 to 15 pages with 3 to 4 subdocuments, enabling evaluation across different complexity scales. Poly variants (PolySeq, PolyInt, PolyRand) incorporate 7 to 8 document types per packet in Large configurations and 3.5 to 3.8 types in Small configurations, testing cross-type boundary detection. Mono variants (MonoSeq, MonoRand) contain exactly one document type per packet (Types/Packet = 1.0), isolating intra-type boundary detection from classification. The pages per subdocument ratio indicates that Mono variants contain longer individual documents (8 to 16 pages) compared to Poly variants (5 to 8 pages), reflecting different real-world scenarios where homogeneous document batches often contain lengthier materials.}
\label{tab:packet-stats}
\begin{tabular}{llllllll}
\toprule
\textbf{Dataset Type} & \textbf{Size} & \textbf{Split} & \textbf{Total} & \textbf{Pages/} & \textbf{Subdocs/} & \textbf{Types/} & \textbf{Pages/} \\
 & & & \textbf{Packets} & \textbf{Packet} & \textbf{Packet} & \textbf{Packet} & \textbf{Subdoc} \\
\midrule
\multirow{6}{*}{PolySeq} 
 & \multirow{3}{*}{Large} & Test & 500 & 69.86 & 12.31 & 7.97 & 5.67 \\
 &  & Train & 1864 & 127.05 & 16.04 & 8.77 & 7.92 \\
 &  & Validation & 199 & 79.51 & 11.10 & 7.50 & 7.16 \\ \cline{2-8}
 & \multirow{3}{*}{Small} & Test & 500 & 14.66 & 4.05 & 3.66 & 3.62 \\ 
 &  & Train & 7840 & 14.89 & 4.31 & 3.85 & 3.46 \\
 &  & Validation & 1834 & 15.16 & 4.15 & 3.78 & 3.65 \\
\midrule
\multirow{6}{*}{PolyInt} 
 & \multirow{3}{*}{Large} & Test & 472 & 69.58 & 12.81 & 7.94 & 5.43 \\
 &  & Train & 1745 & 127.61 & 16.53 & 8.51 & 7.72 \\
 &  & Validation & 190 & 75.30 & 10.79 & 7.08 & 6.98 \\ \cline{2-8}
 & \multirow{3}{*}{Small} & Test & 500 & 14.52 & 4.06 & 3.53 & 3.57 \\ 
 &  & Train & 7728 & 14.82 & 4.33 & 3.70 & 3.42 \\
 &  & Validation & 1776 & 15.16 & 4.14 & 3.60 & 3.66 \\
\midrule
\multirow{6}{*}{PolyRand} 
 & \multirow{3}{*}{Large} & Test & 486 & 69.18 & 12.90 & 7.87 & 5.36 \\
 &  & Train & 1771 & 126.91 & 15.85 & 8.41 & 8.01 \\
 &  & Validation & 161 & 74.47 & 10.14 & 6.74 & 7.34 \\ \cline{2-8}
 & \multirow{3}{*}{Small} & Test & 500 & 14.46 & 4.11 & 3.55 & 3.51 \\ 
 &  & Train & 7756 & 14.75 & 4.30 & 3.66 & 3.43 \\
 &  & Validation & 1807 & 15.19 & 4.22 & 3.64 & 3.60 \\
\midrule
\multirow{6}{*}{MonoSeq} 
 & \multirow{3}{*}{Large} & Test & 102 & 54.89 & 5.43 & 1.00 & 10.11 \\
 &  & Train & 485 & 113.24 & 7.16 & 1.00 & 15.80 \\
 &  & Validation & 45 & 56.62 & 4.58 & 1.00 & 12.37 \\ \cline{2-8}
 & \multirow{3}{*}{Small} & Test & 500 & 13.31 & 3.37 & 1.00 & 3.95 \\
 &  & Train & 5305 & 13.41 & 4.02 & 1.00 & 3.33 \\
 &  & Validation & 966 & 13.36 & 3.28 & 1.00 & 4.07 \\
\midrule
\multirow{6}{*}{MonoRand} 
 & \multirow{3}{*}{Large} & Test & 96 & 45.31 & 5.34 & 1.00 & 8.48 \\
 &  & Train & 417 & 84.84 & 7.03 & 1.00 & 12.06 \\
 &  & Validation & 51 & 42.84 & 4.25 & 1.00 & 10.07 \\ \cline{2-8}
 & \multirow{3}{*}{Small} & Test & 500 & 11.36 & 3.42 & 1.00 & 3.32 \\
 &  & Train & 5471 & 11.60 & 3.97 & 1.00 & 2.92 \\
 &  & Validation & 1052 & 11.10 & 3.17 & 1.00 & 3.50 \\
\bottomrule
\end{tabular}
\end{table}

\begin{table}[!t]
\centering
\caption{Document type distribution across DocSplit benchmark variants showing counts for 13 categories: Budget, Email, Form, Handwritten, Language, Invoice, Letter, News Article, Questionnaire, Memo, Resume, Scientific Publication, and Specification. Poly variants (PolySeq, PolyInt, PolyRand) exhibit balanced representation across all document types, with each category contributing 100 to 350 instances in Large test splits, ensuring unbiased evaluation across diverse formats. Mono variants (MonoSeq, MonoRand) exclude the Language category due to limited source samples, as evidenced by zero or near-zero counts. The stratified splitting maintains proportional type distributions across train, validation, and test partitions, with Form, Letter, and Resume being most prevalent in Mono variants. This distribution enables comprehensive evaluation of both cross-type discrimination in Poly benchmarks and same-type boundary detection in Mono benchmarks.}
\label{tab:doctype-dist}
\small
\begin{tabular}{l|l|ccccccccccccc}
\hline
\textbf{Size} & \textbf{Split} & \rotatebox{90}{\textbf{Budget}} & \rotatebox{90}{\textbf{Email}} & \rotatebox{90}{\textbf{Form}} & \rotatebox{90}{\textbf{Handwritten}} & \rotatebox{90}{\textbf{Invoice}} & \rotatebox{90}{\textbf{Language}} & \rotatebox{90}{\textbf{Letter}} & \rotatebox{90}{\textbf{Memo}} & \rotatebox{90}{\textbf{News Article}} & \rotatebox{90}{\textbf{Questionnaire}} & \rotatebox{90}{\textbf{Resume}} & \rotatebox{90}{\textbf{Sci. Pub.}} & \rotatebox{90}{\textbf{Spec.}} \\
\hline
\multicolumn{15}{c}{\textbf{\dsPolySeq}} \\
\hline
\multirow{3}{*}{Large} & Test & 323 & 288 & 330 & 322 & 314 & 292 & 301 & 322 & 281 & 294 & 287 & 341 & 292 \\
 & Train & 1240 & 1225 & 1361 & 1342 & 1196 & 1224 & 1232 & 1215 & 1246 & 1265 & 1236 & 1359 & 1212 \\
 & Validation & 135 & 98 & 118 & 135 & 109 & 100 & 112 & 113 & 111 & 98 & 119 & 149 & 95 \\
\cline{1-15}
\multirow{3}{*}{Small} & Test & 129 & 139 & 136 & 138 & 135 & 130 & 123 & 135 & 138 & 136 & 160 & 167 & 165 \\
 & Train & 2346 & 2246 & 2045 & 2207 & 2344 & 2683 & 2316 & 2317 & 2393 & 2322 & 2371 & 2231 & 2334 \\
 & Validation & 506 & 530 & 456 & 546 & 523 & 455 & 560 & 579 & 545 & 555 & 504 & 627 & 542 \\
\hline
\multicolumn{15}{c}{\textbf{\dsPolyInt}} \\
\hline
\multirow{3}{*}{Large} & Test & 288 & 262 & 308 & 302 & 271 & 266 & 278 & 286 & 279 & 287 & 293 & 350 & 280 \\
 & Train & 1165 & 1113 & 1226 & 1227 & 1094 & 1074 & 1124 & 1131 & 1123 & 1127 & 1144 & 1222 & 1084 \\
 & Validation & 107 & 92 & 98 & 96 & 104 & 81 & 105 & 111 & 109 & 107 & 115 & 126 & 94 \\
\cline{1-15}
\multirow{3}{*}{Small} & Test & 128 & 133 & 130 & 120 & 129 & 125 & 131 & 134 & 135 & 133 & 133 & 162 & 171 \\
 & Train & 2180 & 2171 & 1921 & 2139 & 2178 & 2431 & 2224 & 2222 & 2322 & 2226 & 2266 & 2068 & 2207 \\
 & Validation & 472 & 514 & 465 & 498 & 502 & 440 & 492 & 508 & 466 & 492 & 491 & 563 & 483 \\
\hline
\multicolumn{15}{c}{\textbf{\dsPolyRand}} \\
\hline
\multirow{3}{*}{Large} & Test & 307 & 263 & 292 & 306 & 278 & 265 & 286 & 307 & 291 & 309 & 292 & 355 & 275 \\
 & Train & 1188 & 1142 & 1229 & 1246 & 1129 & 1090 & 1096 & 1101 & 1140 & 1108 & 1117 & 1196 & 1112 \\
 & Validation & 80 & 94 & 82 & 94 & 81 & 67 & 84 & 102 & 75 & 73 & 83 & 98 & 72 \\
\cline{1-15}
\multirow{3}{*}{Small} & Test & 126 & 134 & 130 & 128 & 133 & 131 & 131 & 128 & 128 & 146 & 142 & 148 & 168 \\
 & Train & 2168 & 2107 & 1982 & 2048 & 2165 & 2390 & 2256 & 2254 & 2294 & 2214 & 2248 & 2092 & 2167 \\
 & Validation & 495 & 490 & 463 & 490 & 520 & 450 & 533 & 546 & 488 & 527 & 490 & 573 & 511 \\
\hline
\multicolumn{15}{c}{\textbf{\dsMonoSeq}} \\
\hline
\multirow{3}{*}{Large} & Test & 12 & 0 & 9 & 39 & 0 & 0 & 5 & 4 & 0 & 1 & 9 & 23 & 0 \\
 & Train & 46 & 12 & 99 & 188 & 0 & 3 & 24 & 12 & 8 & 15 & 14 & 64 & 0 \\
 & Validation & 3 & 0 & 3 & 9 & 0 & 0 & 0 & 4 & 6 & 1 & 5 & 14 & 0 \\
\cline{1-15}
\multirow{3}{*}{Small} & Test & 38 & 6 & 42 & 59 & 35 & 0 & 53 & 55 & 13 & 47 & 53 & 70 & 29 \\
 & Train & 467 & 276 & 353 & 516 & 258 & 443 & 464 & 385 & 391 & 437 & 568 & 546 & 201 \\
 & Validation & 79 & 45 & 18 & 113 & 56 & 0 & 108 & 109 & 46 & 78 & 126 & 137 & 51 \\
\hline
\multicolumn{15}{c}{\textbf{\dsMonoRand}} \\
\hline
\multirow{3}{*}{Large} & Test & 5 & 0 & 13 & 42 & 1 & 0 & 6 & 4 & 0 & 5 & 4 & 16 & 0 \\
 & Train & 42 & 15 & 84 & 139 & 0 & 3 & 20 & 13 & 9 & 18 & 19 & 55 & 0 \\
 & Validation & 3 & 0 & 2 & 13 & 0 & 0 & 2 & 4 & 9 & 1 & 2 & 15 & 0 \\
\cline{1-15}
\multirow{3}{*}{Small} & Test & 47 & 6 & 49 & 56 & 32 & 0 & 47 & 51 & 12 & 50 & 54 & 67 & 29 \\
 & Train & 468 & 278 & 383 & 545 & 219 & 416 & 478 & 452 & 412 & 459 & 567 & 570 & 224 \\
 & Validation & 71 & 53 & 16 & 130 & 61 & 0 & 110 & 125 & 78 & 95 & 108 & 160 & 45 \\
\hline
\end{tabular}
\end{table}

\section{Document Packet Splitting Baseline}
\label{appendix:baseline_approach}

We introduce an LLM-based document splitting approach that decomposes heterogeneous multi-page document packets into constituent subdocuments through structured prompting. The baseline addresses the fundamental challenge where pages from individual documents may be dispersed throughout concatenated packets, requiring simultaneous boundary detection, type classification, and page ordering (Figure~\ref{fig:enter-label}).

LLM-Guided Document Splitting Approach: We design a hierarchical instruction structure comprising three components: system-level role definition establishing the model as a document classification expert, task-specific guidance providing document text through XML-delimited sections with explicit type definitions, and detailed splitting criteria emphasizing content continuity, visual consistency, and logical document structure. The framework explicitly addresses critical challenges including shuffled pages and adjacent documents of identical types requiring separation, while constraining output to predefined document categories through strict type enforcement mechanisms.

Multi-Features Document Processing: To address LLM multimodal input limitations, we implement a text-only processing methodology that maximizes document coverage within context windows. The approach utilizes textual representations extracted via Amazon Textract with LAYOUT, TABLES, FORMS, and SIGNATURES features, converting pages to structured markdown format preserving spatial organization, tabular data, key-value pairs, and signature information. This methodology accommodates arbitrary page counts limited only by model token capacity, leveraging natural language understanding to identify document patterns and structural transitions through rich textual features.

\textbf{Structured Output Generation:} The framework produces hierarchical JSON representations where each subdocument contains document type identification (\texttt{doc\_type\_id}), page ordering through one-based ordinals (\texttt{page\_ordinals}), and unique local identifiers (\texttt{local\_doc\_id}) combining type with ordinal numbering. This structure accommodates complex scenarios where pages from different documents intermix throughout packets. For instance, a 20-page packet containing four distinct documents with shuffled ordering, where scientific publication pages appear at positions $[1, 2, 5, 8, 13, 14, 15, 20]$ while memo pages distribute at $[3, 6, 7, 10, 18]$, requiring simultaneous clustering and sequence reconstruction capabilities. Complete prompt specifications and implementation details appear in Appendix~\ref{appendix:doc_split_prompt}.

\section{Document Splitting Baseline Approach Prompts}
\label{appendix:doc_split_prompt}

\begin{table*}[!h]
\centering
\begin{adjustbox}{width=\textwidth}
\begin{tcolorbox}[
    colback=black!5!white,
    colframe=black,
    colbacktitle=black!15!white,
    coltitle=black,
    fonttitle=\bfseries,
    title=Document Splitting Baseline Approach: Model Configuration,
    halign title=center,
    fontupper=\small,
]

\textbf{Model Configuration Parameters:}

\texttt{temperature: 0.0}\\
\texttt{top\_p: 0.1}\\
\texttt{max\_tokens: 4096}\\
\texttt{top\_k: 5}

\end{tcolorbox}
\end{adjustbox}
\captionof{table}{Model configuration parameters for document splitting baseline approach}
\label{appendix:doc_split_config}
\end{table*}

\begin{table*}[!h]
\centering
\begin{adjustbox}{width=\textwidth}
\begin{tcolorbox}[
    colback=black!5!white,
    colframe=black,
    colbacktitle=black!15!white,
    coltitle=black,
    fonttitle=\bfseries,
    title=Document Splitting Baseline Approach: System Prompt,
    halign title=center,
    fontupper=\small,
]

You are a document classification expert who can analyze and classify multiple documents and their page boundaries within a document package from various domains. Your task is to determine the document type based on its content and structure, using the provided document type definitions. Your output must be valid JSON according to the requested format.

\end{tcolorbox}
\end{adjustbox}
\captionof{table}{System prompt for document splitting baseline approach}
\label{appendix:doc_split_system}
\end{table*}

\begin{table*}[!h]
\centering
\begin{adjustbox}{width=\textwidth}
\begin{tcolorbox}[
    colback=black!5!white,
    colframe=black,
    colbacktitle=black!15!white,
    coltitle=black,
    fonttitle=\bfseries,
    title=Document Splitting Baseline Approach: Task Prompt (Part 1/3),
    halign title=center,
    fontupper=\small,
]

\textbf{\# Document Processing Instructions}

\textbf{\#\# Document Text Structure}

The \verb|<document-text>| XML tags contains the text separated into pages from the document package. Each page will begin with a \verb|<page-number>| XML tag indicating the one based page ordinal of the page text to follow.

\textbf{\#\# Document Types Reference}

The \verb|<document-types>| XML tags contain a markdown table of known doc types for detection.

\textbf{\#\# Terminology Guidance}

\textbf{Guidance for terminology found in the instructions:}
\begin{itemize}
    \item \textbf{ordinal\_start\_page}: The one based beginning page of a document segment within the document package.
    \item \textbf{ordinal\_end\_page}: The one based ending page of a document segment within the document package.
    \item \textbf{document\_type}: The document type code detected for a document segment.
    \item \textbf{Distinct documents} of the same type may be adjacent to each other in the packet. Be sure to separate them into different document segments and don't combine them.
    \item \textbf{Bank Pages} in a document belong to the last document type and are included in the doc\_type page count, for the purposes of ordinal\_end\_page calculation
\end{itemize}

\end{tcolorbox}
\end{adjustbox}
\captionof{table}{Task prompt for document splitting baseline approach - Document structure and terminology}
\label{appendix:doc_split_task_1}
\end{table*}

\begin{table*}[!htbp]
\centering
\begin{adjustbox}{width=\textwidth}
\begin{tcolorbox}[
    colback=black!5!white,
    colframe=black,
    colbacktitle=black!15!white,
    coltitle=black,
    fonttitle=\bfseries,
    title=Document Splitting Baseline Approach: Task Prompt (Part 2/3),
    halign title=center,
    fontupper=\small,
]

\textbf{\#\# Document Splitting Guidance}

When deciding whether pages belong to the same document segment:
\begin{itemize}
    \item \textbf{Content continuity}: Pages with continuing paragraphs, numbered sections, or ongoing narratives likely belong to the same document.
    \item \textbf{Visual/formatting consistency}: Similar layouts, headers, footers, and styling suggest pages belong together.
    \item \textbf{Logical completion}: A document typically has a beginning, middle, and end structure.
    \item \textbf{Document boundaries}: Look for clear indicators of a new document such as new title pages, cover sheets, or significantly different subject matter.
    \item \textbf{Content similarity}: Pages discussing the same topic or subject likely belong to the same document.
    \item \textbf{Shuffled Pages}: Pages in the document packet \textbf{MAY} be shuffled out of order.
    \item \textbf{Document Types}: There may be multiple distinct documents of the same type in a document packet.
\end{itemize}

Pages should be grouped together when they represent a coherent, continuous document, even if they span multiple pages. Split documents only when there is clear evidence that a new, distinct document begins.

\textbf{\#\# CRITICAL INSTRUCTION}

You must \textbf{ONLY} use document types explicitly listed in the \verb|<document-types>| section. Do not create, invent, or use any document type not found in this list. If a document doesn't clearly match any listed type, assign it to the most similar listed type or ``other'' if that option is provided.

\end{tcolorbox}
\end{adjustbox}
\captionof{table}{Task prompt for document splitting baseline approach - Splitting guidance and critical instructions}
\label{appendix:doc_split_task_2}
\end{table*}

\begin{table*}[!htbp]
\centering
\begin{adjustbox}{width=\textwidth}
\begin{tcolorbox}[
    colback=black!5!white,
    colframe=black,
    colbacktitle=black!15!white,
    coltitle=black,
    fonttitle=\bfseries,
    title=Document Splitting Baseline Approach: Task Prompt (Part 3/3),
    halign title=center,
    fontupper=\small,
]

\textbf{\#\# Classification Process}

Follow these steps when classifying documents within the document package:
\begin{itemize}
    \item Analyze the pages in the document packet to identify each as a boundary \texttt{start page} a boundary \texttt{end page}, or a non-boundary \texttt{inner page}.
    \item Identify documents of the same type, that are not the same document but are adjacent to each other in the packet.
    \item Do not combine distinct documents of the same type into a single document segment.
    \item Determine what \verb|<document-types>| each \texttt{start page}, \texttt{end page}, and \texttt{inner page} belongs to. Select ONLY from the \verb|<document-types>|.
    \item Beginning with the \texttt{start page} iterate over each unclassified inner page in the document packet to find the best sequential match.
    \item Repeat this until all pages are sorted and classified.
    \item Before finalizing, verify that each document type in your response exactly matches one from the \verb|<document-types>| list.
\end{itemize}

\textbf{\#\#\# Local Doc ID}

Follow these steps to formulate the local\_doc\_id:
\begin{itemize}
    \item The local\_doc\_id exists to identify individual instances of a single document type.
    \item The local\_doc\_id format is made up of \{doc\_type\_id\}-\#\# where \#\# is the 01 based position of that doc type in the document packet.
    \item The classification result JSON structure provides an example made up of an invoice, a letter, and a scientific publication.
    \item Assign a unique local\_doc\_id to each subdocument classified in the document packet, according to the local\_doc\_id format.
\end{itemize}

\end{tcolorbox}
\end{adjustbox}
\captionof{table}{Task prompt for document splitting baseline approach - Classification process and local document ID}
\label{appendix:doc_split_task_3}
\end{table*}

\clearpage

\begin{tcblisting}{
    listing engine=listings,
    listing options={
        basicstyle=\small\ttfamily,
        breaklines=true,
        showstringspaces=false
    },
    colback=gray!5,
    colframe=gray!70,
    listing only,
    breakable,
    title={Listing 1: Ground truth data structure for subdocument classification}
}
{
    "subdocuments": [
        {
            "doc_type_id": "invoice",
            "page_ordinals": [1, 4],
            "local_doc_id": "invoice-01",
        },
        {
            "doc_type_id": "letter", 
            "page_ordinals": [3], 
            "local_doc_id": "letter-01"
        },
        {
            "doc_type_id": "scientific publication",
            "page_ordinals": [2, 5, 6, 7, 8, 9, 10, 11, 12, 13, 14],
            "local_doc_id": "scientific publication-01",
        },
        {
            "doc_type_id": "letter", 
            "page_ordinals": [15], 
            "local_doc_id": "letter-02"
        },
    ]
}
\end{tcblisting}

.
\begin{figure}[!ht]
    \centering
    \includegraphics[width=\linewidth]{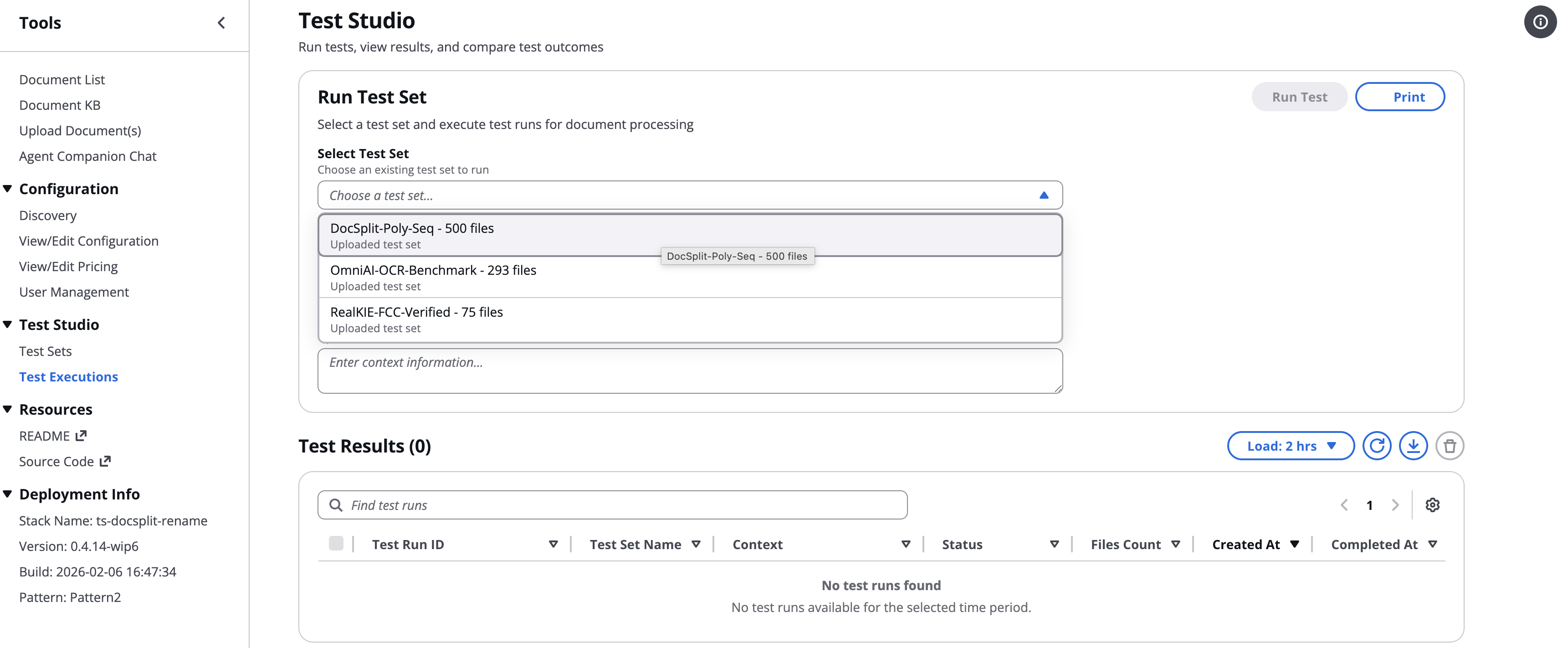}
    \caption{Test Studio interface with pre-populated benchmark datasets for document processing evaluation.}
    \label{fig:test_studio}
\end{figure}

\section{Performing Benchmarking using the IDP Accelerator Test Studio}
\label{appendix:accelerator_baselines}



We developed a web-based evaluation framework\footnote{\url{https://github.com/aws-solutions-library-samples/accelerated-intelligent-document-processing-on-aws/blob/main/docs/test-studio.md}} that enables researchers to benchmark document splitting and classification models without writing code (Figure~\ref{fig:test_studio}). The framework supports both packet splitting metrics and key information extraction metrics, and users may define custom metrics as needed for other downstream tasks. The framework provides a graphical interface for executing batch evaluations on test sets, monitoring progress in real-time, and analyzing results through interactive dashboards. The DocSplit dataset is included by default alongside a handful of other IDP related datasets, and researchers can also import their own custom datasets as described later in this appendix. The evaluation framework is part of the GenAI IDP Accelerator,\footnote{\url{https://github.com/aws-solutions-library-samples/accelerated-intelligent-document-processing-on-aws}} an open-source solution deployed via AWS CloudFormation. For deployment instructions, see the Deployment Guide.\footnote{\url{https://github.com/aws-solutions-library-samples/accelerated-intelligent-document-processing-on-aws/blob/main/docs/deployment.md}}

The framework includes a configuration editor\footnote{\url{https://github.com/aws-solutions-library-samples/accelerated-intelligent-document-processing-on-aws/blob/main/docs/configuration.md}} that allows users to make fine-grained changes to their document processing pipeline, such as adjusting prompts, classification thresholds, or model parameters. Users may even leverage customized fine-tuned models for stages of the processing.  After modifying configurations, users can run evaluations and compare results side-by-side to assess the impact of each change.

The framework supports the complete evaluation workflow: users select a test set containing packet documents with ground truth annotations, configure optional parameters such as file limits for quick iteration, and initiate batch processing. The system automatically processes each packet through the document splitting pipeline, compares predictions against ground truth, and calculates comprehensive metrics. Results are displayed in an interactive dashboard and can be exported in JSON or CSV formats for further analysis.

To execute an evaluation, users select a test set from the available datasets and optionally specify a file limit to process a subset of packets for rapid prototyping. A context field allows users to annotate test runs with descriptions such as ``baseline model'' or ``improved splitting algorithm'' for later reference.

\begin{figure}[!t]
    \centering
    \includegraphics[width=\linewidth]{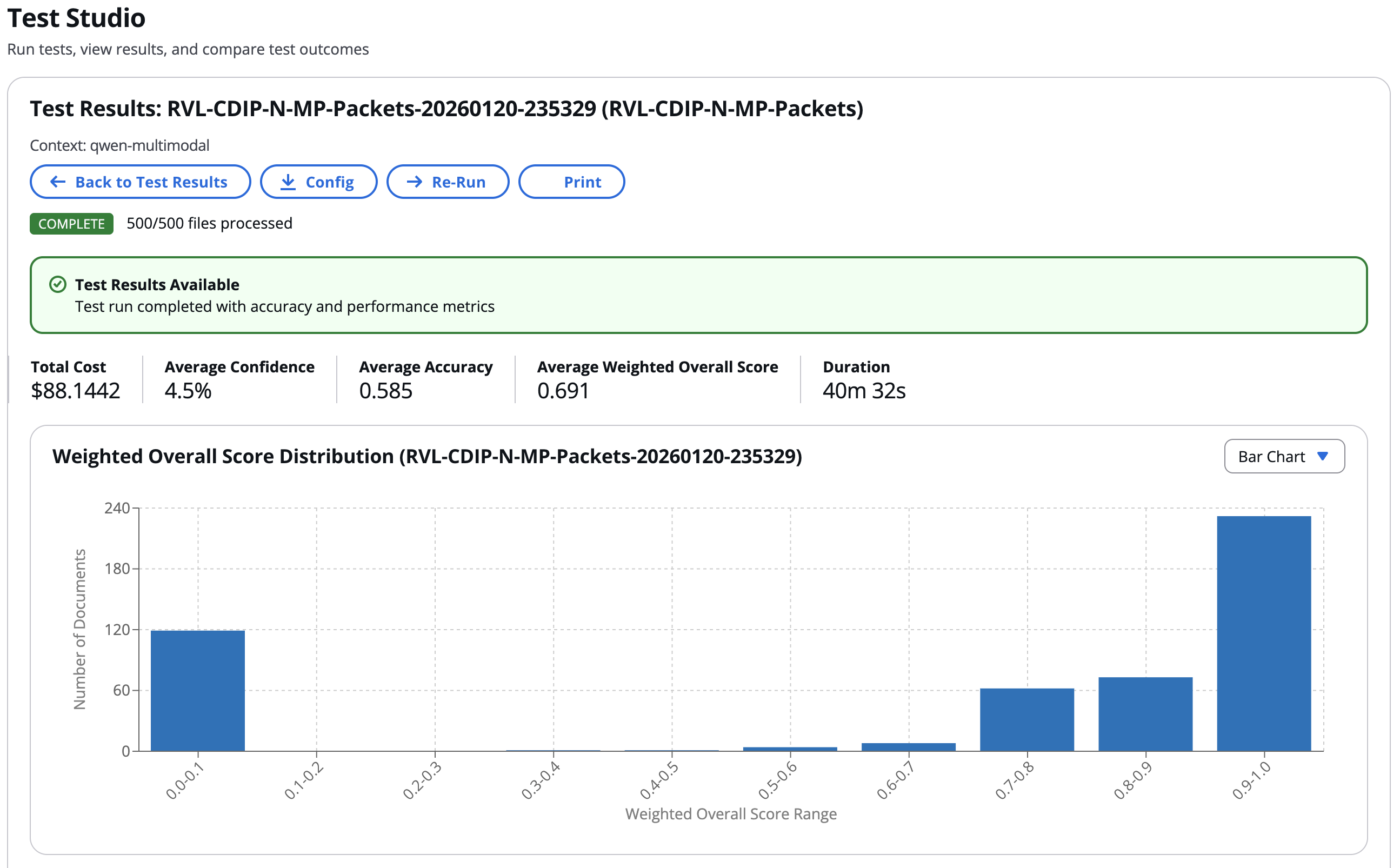}
    \caption{Test Studio evaluation dashboard displaying benchmark results with performance metrics and score distribution.}
    \label{fig:evaluation_report}
\end{figure}

\begin{figure}[!h]
    \centering
    \includegraphics[width=\linewidth]{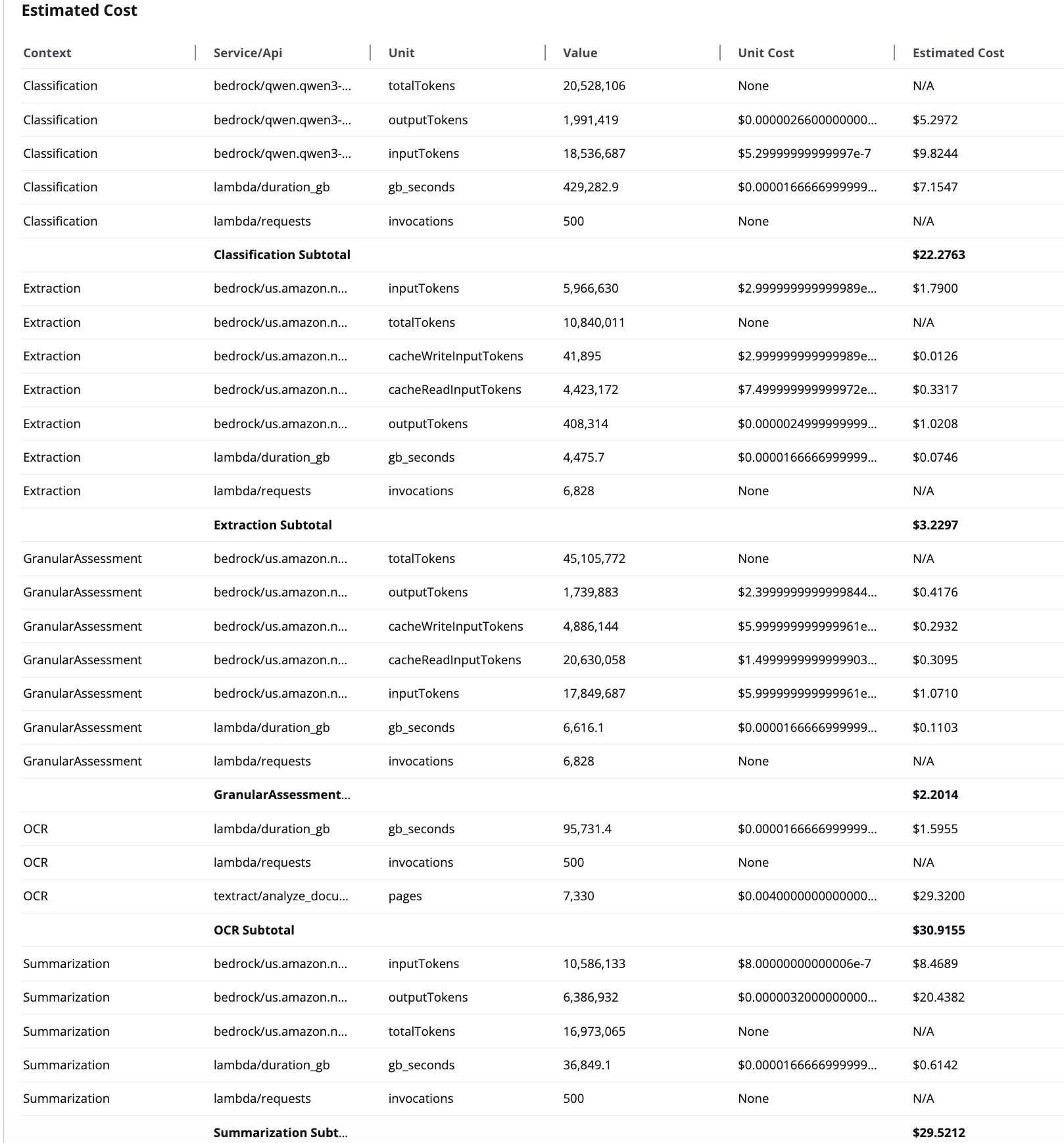}
    \caption{Test Studio cost estimation interface displaying itemized costs across pipeline stages (Classification, Extraction, OCR, Granular Assessment, Summarization) with per-API token and invocation tracking.}
    \label{fig:cost_breakdown}
\end{figure}

Upon initiating a test run, the framework queues all packet files for processing and displays real-time progress including the number of files processed, elapsed time, and current status. When processing completes, results appear in a historical list showing timestamp, test set name, context description, file count, and detailed evaluation report (Figure~\ref{fig:evaluation_report}). The framework's itemized cost attribution enables systematic cost-performance ablation studies, allowing researchers to identify optimal configurations that balance computational expense against splitting accuracy (Figure~\ref{fig:cost_breakdown}). This capability is particularly valuable for industry-specific deployments, such as high-throughput claims processing in healthcare, compliance-critical document handling in finance, or cost-sensitive logistics operations, where accuracy requirements and budget constraints vary significantly. The framework also enables users to inspect individual low-performing documents to diagnose the source of failures.



\subsection{Importing Custom Datasets}
\begin{figure}
    \centering
    \includegraphics[width=0.6\linewidth]{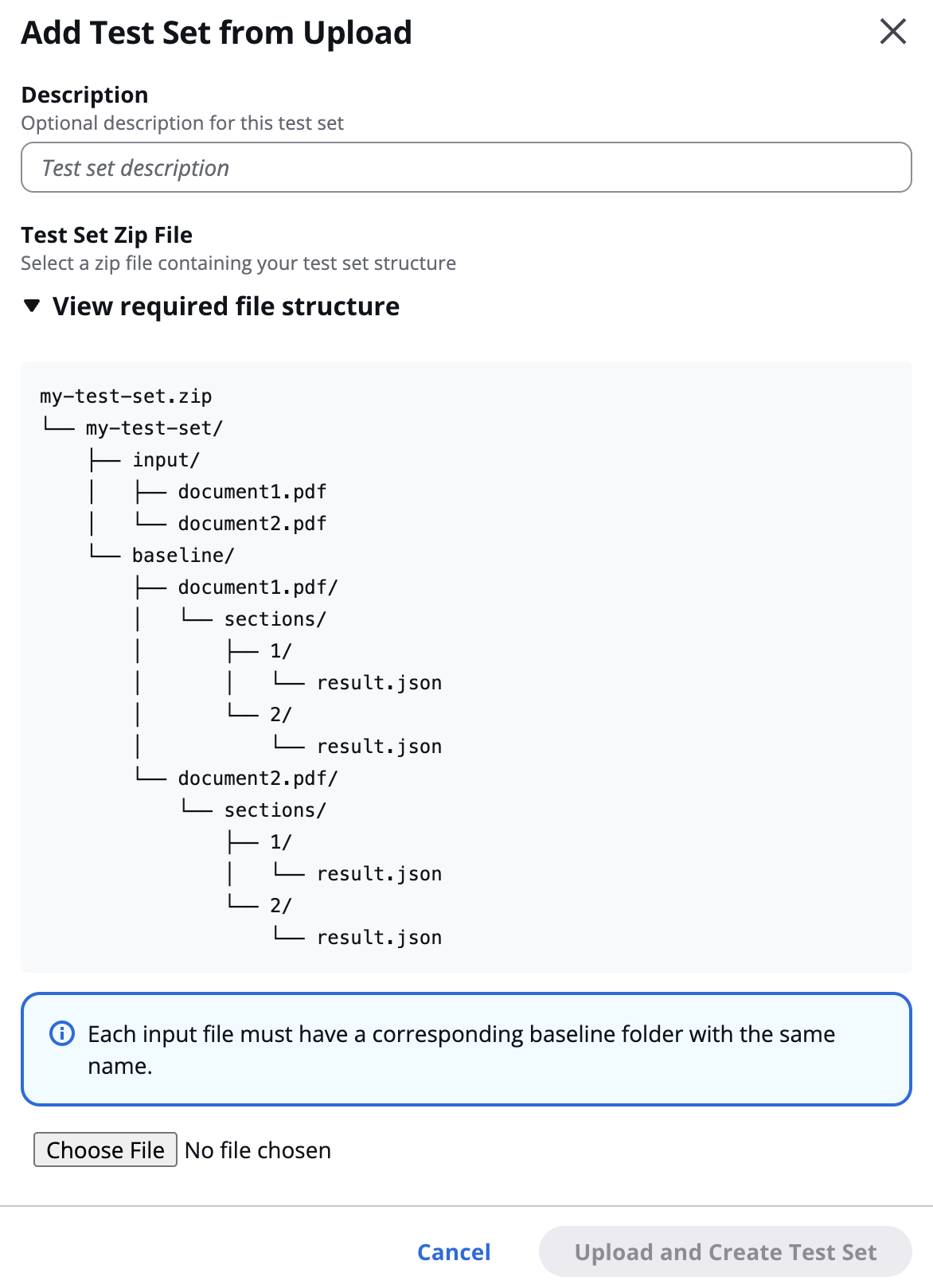}
    \caption{Custom dataset import interface displaying the required structure with input/ folder for packet PDFs and baseline/ folder containing ground truth annotations in result.json files.}
    \label{fig:custom_dataset_format}
\end{figure}

Researchers can import custom packet splitting datasets to evaluate their own document collections. The framework accepts datasets as ZIP archives containing the \texttt{input} and \texttt{baseline} folders at the root level (Figure~\ref{fig:custom_dataset_format}). For comprehensive details on baseline data formats and evaluation methods, see the Evaluation Framework documentation.\footnote{\url{https://github.com/aws-solutions-library-samples/accelerated-intelligent-document-processing-on-aws/blob/main/docs/evaluation.md\#data-structure-requirements}}


Upon upload, the framework validates the directory structure, ensuring each input file has a corresponding baseline folder with properly formatted \texttt{result.json} files. Validation failures report specific issues such as missing baseline folders or malformed JSON. For larger datasets, direct upload to cloud storage is also supported.

Before running evaluations on custom datasets, users should verify that the evaluation configuration includes class definitions matching the document types in their ground truth annotations. Class names in the configuration must exactly match the \texttt{document\_class.type} values in the annotation files. For detailed guidance on configuring document classes, see the Configuration documentation.\footnote{\url{https://github.com/aws-solutions-library-samples/accelerated-intelligent-document-processing-on-aws/blob/main/docs/configuration.md}}

\subsubsection{Ground Truth Data Format.}
The evaluation framework expects ground truth annotations in a specific directory structure. Each test set contains two top-level folders: an \texttt{input} folder with packet PDF files, and a \texttt{baseline} folder with corresponding annotations.

\begin{minipage}{\linewidth}
Ground truth dataset directory structure:

\begin{verbatim}
test-set/
|-- input/
|   |-- packet_001.pdf
|   |-- packet_002.pdf
|   +-- ...
+-- baseline/
    |-- packet_001.pdf/
    |   +-- sections/
    |       |-- 1/
    |       |   +-- result.json
    |       |-- 2/
    |       |   +-- result.json
    |       +-- ...
    +-- ...
\end{verbatim}
\end{minipage}

Each packet PDF in the \texttt{input} folder must have a corresponding folder in \texttt{baseline} with the exact same filename (including the \texttt{.pdf} extension). Within each baseline folder, a \texttt{sections} subfolder contains numbered directories (1, 2, 3, etc.) representing each document section within the packet. Each numbered directory contains a \texttt{result.json} file specifying the ground truth for that section.

\subsubsection{Annotation Schema}
Each \texttt{result.json} file defines the ground truth for a single document section using the following schema:

\begin{verbatim}
{
    "document_class": {
        "type": "<class_name>"
    },
    "split_document": {
        "page_indices": [<zero-indexed page numbers>]
    },
    "inference_result": {}
}
\end{verbatim}

The \texttt{document\_class.type} field specifies the document class as a string, using underscores instead of spaces (e.g., \texttt{news\_article} rather than \texttt{news article}). The \texttt{split\_document.page\_indices} field is an array of zero-indexed page numbers belonging to this section. For a 3-page invoice starting on the first page of a packet, the indices would be \texttt{[0, 1, 2]}. The \texttt{inference\_result} field can remain empty for packet splitting evaluation.

\subsubsection{Annotation Examples}
The following examples illustrate annotations for sections of varying length from packet\_0001 in the RVL-CDIP-N-MP-Packets dataset, a 20-page packet containing 6 distinct documents.

\begin{minipage}{\linewidth}

A single-page section (sections/1/result.json --- Invoice on page 0):
\begin{verbatim}
{
    "document_class": { "type": "invoice" },
    "split_document": { "page_indices": [0] },
    "inference_result": {}
}
\end{verbatim}
\end{minipage}

A multi-page section (sections/3/result.json --- Language document spanning pages 2--4):
\begin{verbatim}
{
    "document_class": { "type": "language" },
    "split_document": { "page_indices": [2, 3, 4] },
    "inference_result": {}
}
\end{verbatim}

A long section (sections/6/result.json --- Form spanning pages 8--19):
\begin{verbatim}
{
    "document_class": { "type": "form" },
    "split_document": { "page_indices": [8, 9, 10, 11, 12, 13, 14, 15, 16, 17, 18, 19] },
    "inference_result": {}
}
\end{verbatim}

Page indices are zero-indexed and contiguous within each section. Section numbers in folder names (1, 2, 3, etc.) correspond to document order within the packet.

\subsubsection{Classification Methods Supported by Test Studio}

The Test Studio allows users to benchmark and compare different classification methods\footnote{\url{https://github.com/aws-solutions-library-samples/accelerated-intelligent-document-processing-on-aws/blob/main/docs/classification.md}} available in the IDP Accelerator. Each method offers distinct trade-offs in accuracy, speed, cost, and configurability. The following subsections describe each supported classification approach.

\paragraph{Text-Based Holistic Classification (LLM-Based).}
Text-Based Holistic Classification analyzes entire document packets to identify logical boundaries and classify document segments in a single LLM request. The model receives the full document text and uses content continuity, formatting consistency, logical structure, and boundary indicators to determine where one document ends and another begins. It then produces a structured output identifying each subdocument with its type and page range.

This method is best suited when documents have clear logical boundaries based on content, when text context spans multiple pages requiring full-document understanding, and when high-context models are available. It works well when visual elements are less important than textual continuity and when document packets fit within the model's context window. However, long documents can exceed the context window of smaller models, lengthy inputs may dilute the model's focus leading to inconsistent classifications, and it is not ideal for very large or visually complex document sets.

\paragraph{MultiModal Page-Level Classification with Sequence Segmentation (LLM-Based, Default).}
This is the default classification method in the IDP Accelerator. It classifies each page independently using both text and image data, and implements a sophisticated sequence segmentation approach similar to BIO (Begin-Inside-Outside) tagging commonly used in NLP. Each page receives a document type label (e.g., ``invoice'' or ``letter'') and a boundary indicator: ``start'' to signal the beginning of a new document or ``continue'' to indicate the current document is ongoing.

This approach automatically segments multi-document packets, even when consecutive documents share the same type. For example, in a packet containing two invoices and one letter, the system correctly creates three separate sections by detecting that the second invoice starts a new document despite having the same classification as the first. This method is best suited when packets contain multiple documents of the same type, when visual layout and image content are important, when processing very large packets that might exceed context limits, and when processing speed matters since parallel page processing is possible. It supports optional few-shot examples, configurable page context for improved boundary detection, and precise control over image placement in prompts. There is no practical limit on document length.

\paragraph{Regex-Based Lightweight Classification (Non-LLM).}
The IDP Accelerator supports optional regex-based classification that bypasses LLM calls entirely when document patterns are recognized through regular expression matching. Two strategies are available: \textit{document name regex}, which classifies all pages based on filename or document ID, and \textit{page content regex}, which classifies individual pages based on OCR text patterns with automatic LLM fallback for unmatched pages. This approach achieves roughly 100 to 1000 times faster processing than LLM classification with zero token usage for matched documents, and includes robust error handling that gracefully falls back to LLM when patterns are invalid or fail at runtime.

However, regex-based classification has important limitations. Patterns are inherently tied to text patterns observed in the development dataset and may \textbf{overfit}, failing to generalize to unseen documents with different formatting or terminology. Regex operates on surface-level text matching without semantic understanding, meaning a document merely mentioning ``invoice'' in its body could be misclassified. Small formatting variations can cause mismatches, and new document variations may require continuous pattern updates. Despite these limitations, this approach is an excellent choice when documents follow well-known, predictable patterns with consistent formatting, when speed and cost are critical, or as a fast pre-filter with LLM fallback for ambiguous cases.

\section{Limitations}
\textbf{Long Document Packets}: While {\ds} establishes a comprehensive foundation for document packet splitting evaluation, several aspects warrant further investigation in future work. Our current benchmark focuses on packets with moderate page counts, typically ranging from 5 to 20 pages per packet. Future iterations could explore scenarios involving substantially longer document packets (e.g., 100+ pages) to assess scalability and performance degradation patterns in extended contexts, which would provide valuable insights for enterprise-scale document processing systems.

\textbf{Model Diversity}: The present study evaluates Claude Sonnet 4.5, Claude Haiku 4.5, DeepSeek, Gemma, and Qwen as our primary baselines, deliberately chosen to demonstrate the efficacy of our proposed evaluation framework and benchmark datasets. This focused approach allows us to establish clear performance baselines and validate our metric design. However, the framework's generalizability across diverse model architectures remains an open question. Future research directions include systematic evaluation of additional large language models with varying architectural paradigms, parameter scales, and training objectives to comprehensively characterize the document packet splitting task space.

\textbf{Multimodal Features}: Our current experimental design prioritizes text-based document understanding, leveraging rich textual features extracted via AWS Textract. While this approach enables processing of arbitrary-length documents within token constraints, it does not fully exploit visual and layout information that vision-language models (VLMs) can leverage. Future work could investigate multimodal baselines that jointly process visual and textual modalities, potentially revealing complementary strengths in boundary detection and page ordering tasks. Such investigations would be particularly valuable for documents where visual layout provides critical structural cues.

\textbf{Evaluation Metrics}: The evaluation framework introduced in this work combines established clustering metrics (Rand Index, V-measure) with ordering correlation measures (Kendall's Tau). While these metrics effectively capture the dual objectives of document clustering and page sequencing, alternative formulations merit exploration. Future research could investigate additional clustering algorithms, distance metrics, and ordering measures to develop more robust and nuanced evaluation protocols. Furthermore, exploring document splitting strategies that incorporate hierarchical document structure or leverage domain-specific priors could yield performance improvements in specialized application contexts.


\end{document}